\pdfoutput=1

\documentclass[11pt]{article}

\usepackage{acl}

\usepackage{times}
\usepackage{latexsym}

\usepackage[T1]{fontenc}

\usepackage[utf8]{inputenc}

\usepackage{microtype}

\usepackage{inconsolata}

\usepackage{subcaption}
\usepackage{booktabs}
\usepackage{graphicx}
\usepackage{enumitem}
\usepackage{amsmath} 
\usepackage{listings}
\usepackage{adjustbox}

\usepackage{colortbl}

\usepackage{multirow}
\usepackage{xcolor}
\usepackage{hyperref}
\usepackage{xspace}

\addtocontents{toc}{\protect\setcounter{tocdepth}{-1}}

\newcommand*\samethanks[1][\value{footnote}]{\footnotemark[#1]}

\definecolor{darkgreen}{RGB}{50,100,0}
\definecolor{darkred}{RGB}{200, 0, 0}
\definecolor{lightred}{RGB}{250, 200, 200}
\definecolor{lightblue}{RGB}{200, 200, 250}

\definecolor{pink}{rgb}{0.858, 0.188, 0.478}  
\hypersetup{colorlinks=true, urlcolor=pink}

\newcommand{\ie}{\emph{i.e.,}\xspace}
\newcommand{\eg}{\emph{e.g.,}\xspace}

\newcommand{\model}[1]{\textsc{CriticBench}} %

\lstdefinestyle{python}{
    language=Python,
    basicstyle=\fontsize{7}{9.5}\ttfamily,
    keywordstyle=\color{blue},
    commentstyle=\color{gray},
    stringstyle=\color{black},
    showstringspaces=false,
    breaklines=true,
    breakindent=0pt,
    breakatwhitespace=false,
    escapeinside={(*@}{@*)}
}

\title{\model{}: Benchmarking LLMs for Critique-Correct Reasoning}

\author{Zicheng Lin$^{1}$\thanks{~Equal contribution.}, ~Zhibin Gou$^{1}$\samethanks, ~Tian Liang$^{1}$, ~Ruilin Luo$^{1}$, ~Haowei Liu$^{2}$, ~Yujiu Yang$^{1}$\thanks{~Corresponding author.}\\
   $^{1}$Tsinghua University
   $^{2}$University of Hong Kong \\
  \texttt{linzc23@mails.tsinghua.edu.cn, zebgou@gmail.com}\\
\url{https://criticbench.github.io}}

\begin{document}
\maketitle

\begin{abstract}
The ability of Large Language Models (LLMs) to critique and refine their reasoning is crucial for their application in evaluation, feedback provision, and self-improvement. This paper introduces \model{}, a comprehensive benchmark designed to assess LLMs' abilities to critique and rectify their reasoning across a variety of tasks.
\model{} encompasses five reasoning domains: mathematical, commonsense, symbolic, coding, and algorithmic.
It compiles 15 datasets and incorporates responses from three LLM families.
Utilizing \model{}, we evaluate and dissect the performance of 17 LLMs in generation, critique, and correction reasoning, \ie GQC reasoning, and analyze the key factors affecting LLM critical reasoning.
Our findings reveal: (1) a linear relationship in GQC capabilities, with critique-focused training markedly enhancing performance; (2) a task-dependent variation in critique and correction effectiveness, with logic-oriented tasks being more amenable to correction; (3) GQC knowledge inconsistencies that decrease as model size increases; and
(4) an intriguing inter-model critiquing pattern, where stronger models are better at critiquing weaker ones, while weaker models can surprisingly surpass stronger ones in their self-critique.
We hope these insights into the nuanced critique-correct reasoning of LLMs will foster further research in LLM critique and self-improvement\footnote{~Code and data are available at \url{https://github.com/CriticBench/CriticBench}.}.

\end{abstract}

\begin{figure}[t]
    \centering
    \includegraphics[width=\linewidth]{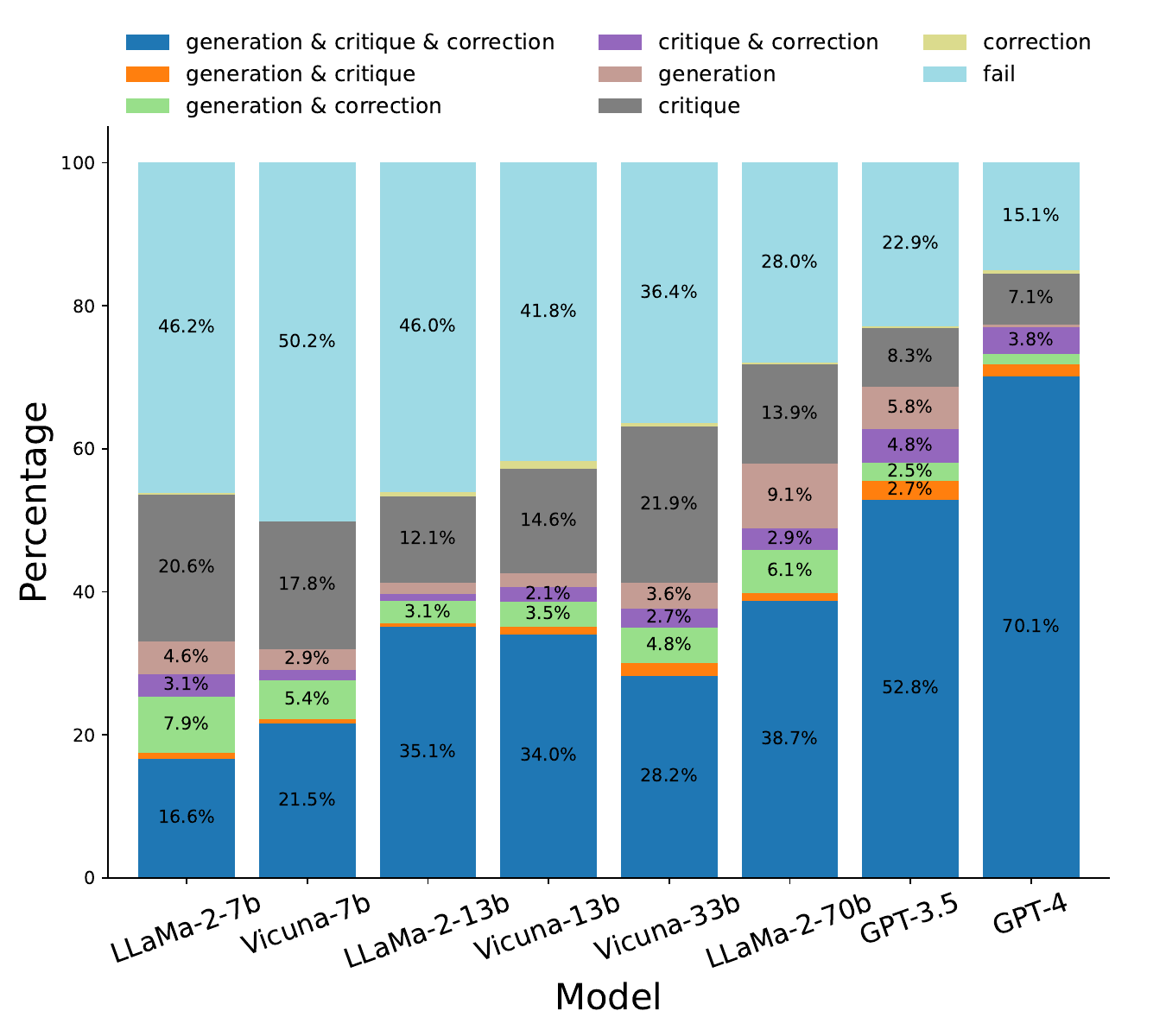}
    \caption{Knowledge consistency across a model's three abilities: generation (G), critique (Q), and correction (C). It shows the overlap of correct responses by the model for a given question across these tasks. }
    \label{fig: g_d_c_distribution}
\end{figure}
\begin{figure*}[t]
    \centering
    \includegraphics[width=0.95\linewidth]{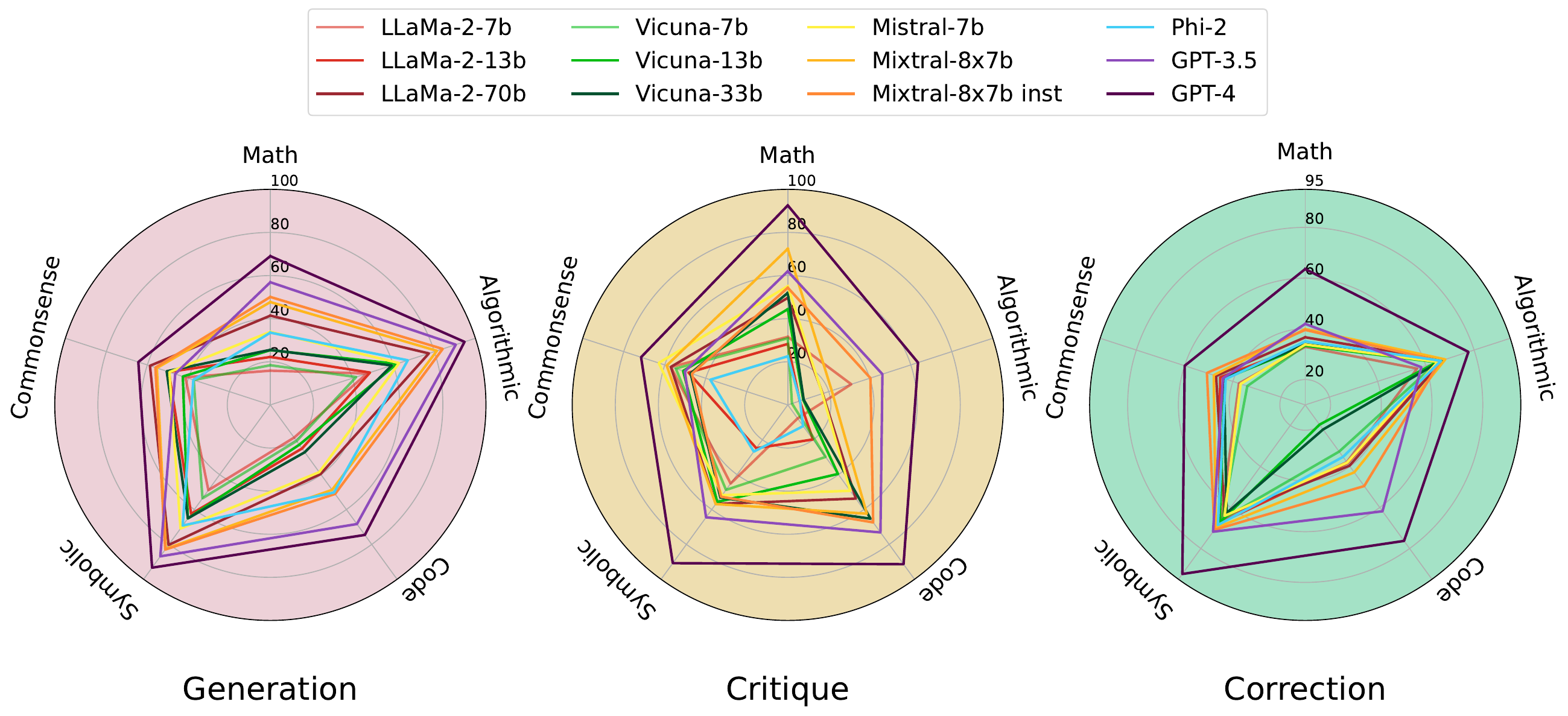}
    \caption{Models' performance on \model{}.}
    \label{fig:rader}
\end{figure*}

\section{Introduction}

The advent of large language models (LLMs) has revolutionized artificial intelligence, showcasing remarkable proficiency across diverse tasks \citep{brown2020language, ouyang2022training, achiam2023gpt, chowdhery2023palm, touvron2023llama,  team2023gemini}. Their potential for self-evaluation and improvement is particularly fascinating, with studies suggesting that LLMs can effectively assess model outputs \citep{liu2023gpteval, fu2023gptscore, chiang2023vicuna}, and even engage in self-reflection and correction \citep{bai2022constitutional, saunders2022self, madaan2023self, gou2024critic}. This capability rests on the LLMs' \emph{critical reasoning} skills, which involves (1) \emph{critique} - identifying issues in provided responses, and (2) \emph{correct} - proposing suitable modifications.

However, a comprehensive understanding of LLMs' critical reasoning abilities remains elusive. Prior research \citep{lightman2023let, li2023generative, luo2023critique} has focused on a narrow range of models and datasets and has yielded inconsistent findings \citep{madaan2023self, huang2023large}, underscoring the need for a thorough investigation. It is imperative to systematically gauge LLMs' proficiency in critiquing and correcting provided answers.

To address these challenges, we present \model{}, a comprehensive benchmark designed to evaluate the critique and correction skills of LLMs. \model{} encompasses 15 datasets spanning five task categories: mathematical, commonsense, symbolic, coding, and algorithmic. We leverage eight models from the LLaMA, Vicuna, and GPT series to create the responses that are to be critiqued and corrected. Moreover, we include GPT-4 and undertake manual data reviews to ensure data quality, culminating in 3.8K data instances (see Table~\ref{tab:dataset_detail} for detailed data collection information).
We conduct extensive experiments on \model{} with 17 LLMs, including closed-source models GPT-3.5 and GPT-4, open-source models like Phi-2 \citep{javaheripi2023phi}, the LLaMA family \citep{touvron2023llama}, the Vicuna family \citep{chiang2023vicuna}, and the Mistral family \citep{jiang2023mistral, jiang2024mixtral}, as well as two models specifically trained for critiquing, namely Auto-J \citep{li2023generative} and UltraCM \citep{cui2023ultrafeedback}. Partial results are shown in Figure \ref{fig:rader}.

\begin{figure*}[t]
    \centering
    \includegraphics[width=0.95\linewidth]{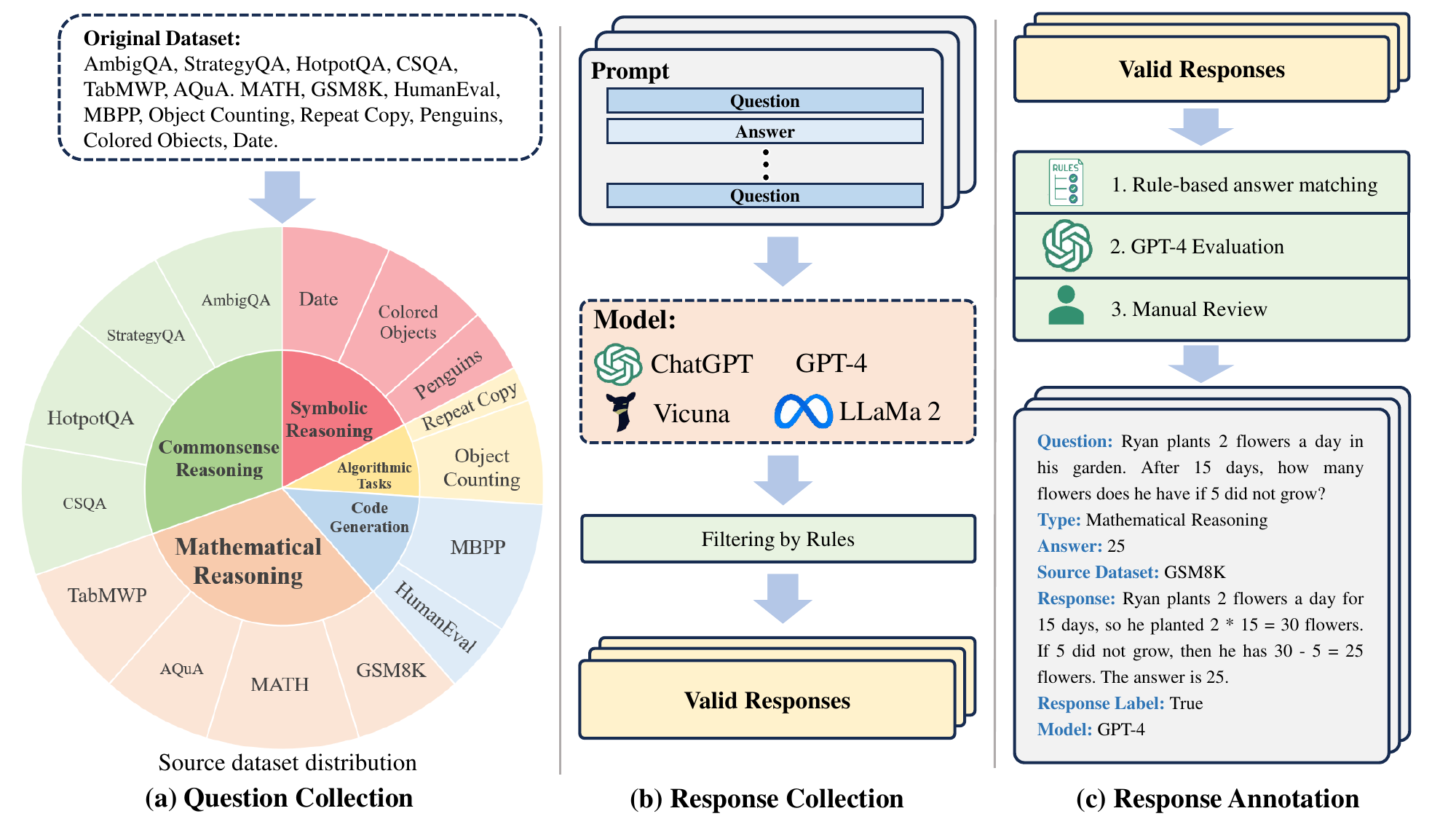}
    \caption{An overview for the \model{} construction.}
    \label{fig: overview}
\end{figure*}
Our contributions are summarized as follows:

\begin{itemize}
\item We present \model{}, a benchmark comprising five different domains to systematically assess critique and correction reasoning in various LLMs. Utilizing \model{}, we investigate the impact of base models, training strategies, prompt strategies, and oracle feedback on the critique-correct reasoning performance of LLMs.

\item We reveal that LLMs exhibit a linear relationship in their generation, critique, and correction (GQC) capabilities, despite limited training on critique tasks.

\item The type of task has a significant impact on correction performance. LLMs struggle more with incorrect answers in detail-oriented tasks like algorithmic tasks compared to logic-centric tasks like code generation.

\item By comparing the sets of questions where models correctly generate, critique, and correct, we find that a model's knowledge is inconsistent across these three tasks, with stronger models showing more consistency in GQC capabilities.

\item We observe that although stronger models are better at critiquing, models with weaker generative abilities can still accurately evaluate responses from stronger models, sometimes even outperforming the latter in self-critique.

\end{itemize}

\section{Related Work}
\paragraph{LLM Reasoning}
The advent of few-shot learning and Chain of Thought (CoT) prompting \cite{brown2020language, wei2022chain} has significantly improved the performance of LLMs in reasoning tasks.
Since then, various advanced prompting methods \citep{wang2022self, zhou2022least, fu2022complexity, zheng2023progressive} have achieved remarkable results, and researchers have proposed to use external tools like search engines \cite{nakano2021webgpt, gou2024critic} and Python interpreters \cite{chen2022program, gao2023pal, gou2024tora} to further augment LLM reasoning in various tasks \cite{schick2024toolformer, ma2024sciagent}.

\paragraph{LLM Critiquing \& Correction}
As LLMs continue to evolve in their capabilities across various domains, many studies seek to enhance performance by requiring LLMs to provide critiques of the generated responses in various forms, including utilizing internal feedback \citep{bai2022constitutional, saunders2022self, welleck2022generating, madaan2023self, zheng2023judging}, leveraging external feedback \cite{kim2023language, shinn2023reflexion, gou2024critic, chen2023teaching}, and employing multiple models for critiquing and debating answers \citep{liang2023encouraging, du2023improving, yin2023exchange}. These studies all demonstrate the potential of LLMs in Critique-Correcting Reasoning. Additionally, some researchers have trained critique models to provide supervisory signals for generative models \citep{ke2023critiquellm, li2023generative, ye2023selfee, wang2023shepherd, cui2023ultrafeedback}. However, these works focus on specific methods without providing a comprehensive evaluation and analysis of Critique-Correcting Reasoning, which also limits the development of this field.

\paragraph{Critiquing Tasks}
Many researchers have advanced the field by creating critiquing datasets, covering areas such as text generation \cite{stiennon2020learning, matiana2021cut}, semantic understanding \cite{pougue2021debagreement}, factuality \cite{thorne2018fever}, alignment \cite{li2023generative}, and mathematics \cite{lightman2023let, luo2023critique}. However, these datasets are typically limited to specific tasks and models. In contrast, our work introduces \model{} to provide the first comprehensive and comparative analysis of LLMs abilities in generation, critique, and correction (GQC).

\section{The \model{}}
\label{sec:criticbench}
\subsection{Overview of \model{}}
\model{} is designed to assess the two key aspects of LLMs' critical reasoning: critique and correction. By combining these two aspects, LLMs can critique a given response and apply corrective reasoning to produce an updated answer. 
In this section, we detail the principles and processes involved in the construction of \model{}; the construction process is listed in Figure \ref{fig: overview}.\\
\model{} is designed to follow these collection principles: (1) it encompasses multiple task types, aimed at comprehensively showcasing the model's abilities; (2) it incorporates diverse models for response generation, promoting response variety; (3) it employs acknowledged datasets, enabling straightforward comparisons with the models' generation capabilities; and (4) it ensures data quality through both GPT-4 and manual review. 
\begin{figure}[t]
    \centering
    \includegraphics[width=0.9\linewidth]{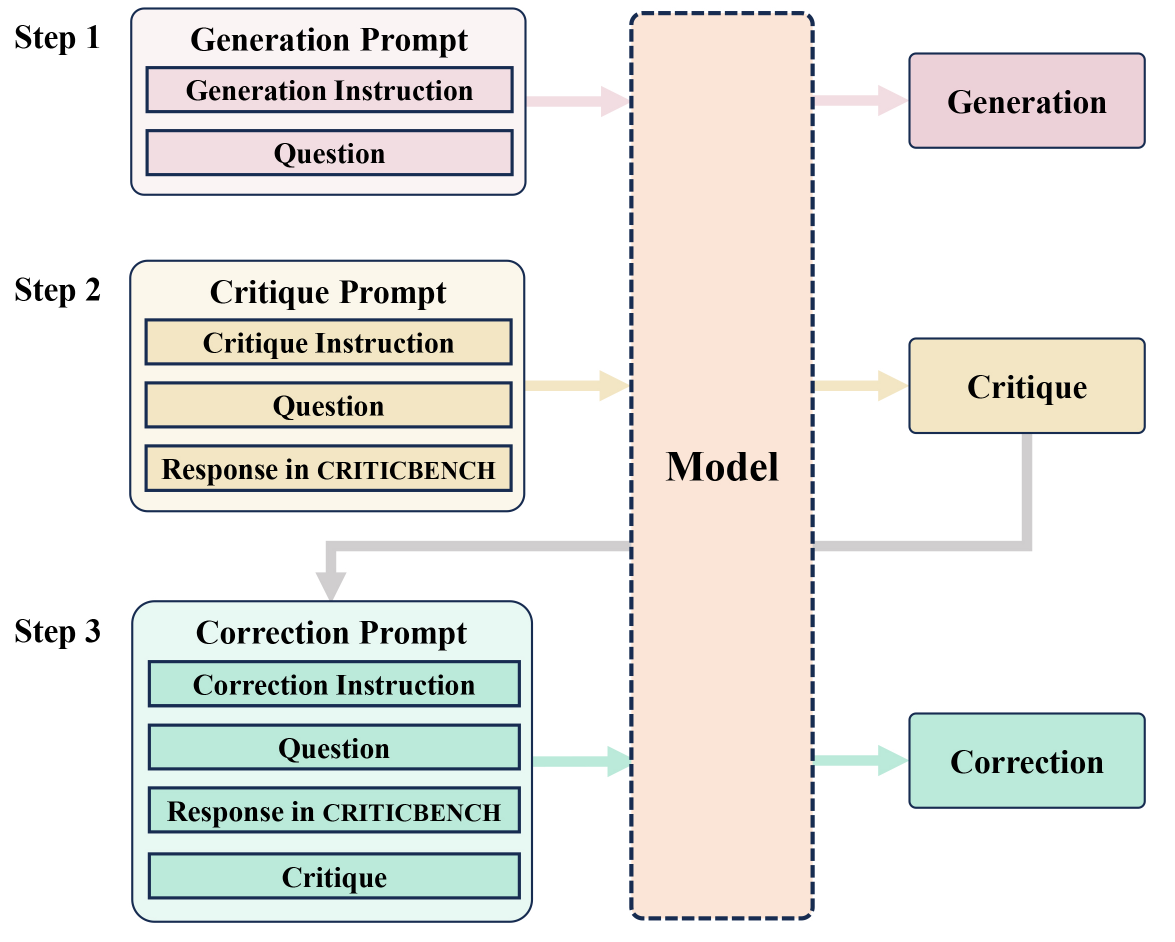}
    \caption{Evaluation process on \model{}.}
    \label{fig:flow}
\end{figure}
\subsection{Question Collection}
This section describes the data collection methodology, following defined principles. A specific quantity of data is extracted from an existing dataset, utilizing any relevant subset if available, or alternatively, selecting randomly from the dataset.\\
\paragraph{Mathematical Reasoning} 
We selected GSM8K \citep{cobbe2021training}, MATH \citep{hendrycksmath2021}, AQuA \cite{ling2017program}, and TabMWP \citep{lu2023dynamic} for mathematical reasoning. We utilized the existing subsets from \citet{shi2022language} and \citet{lightman2023let} respectively, randomly sampled 300 questions from TabMWP, and incorporated all questions from AQuA. \\

\paragraph{Commonsense Reasoning}
To thoroughly evaluate the GQC ability of LLMs in commonsense reasoning tasks, we employ four datasets: CSQA \citep{talmor-etal-2019-commonsenseqa}, AmbigNQ \citep{min2020ambigqa}, StrategyQA \citep{geva2021strategyqa}, and HotpotQA \citep{yang2018hotpotqa}. We randomly sample 300 questions from CSQA, HotpotQA, and AmbigNQ, and include all questions in StrategyQA.

\paragraph{Symbolic Reasoning}
To enrich the variety of symbolic reasoning question types, we utilized three datasets from BIG Bench \cite{srivastava2023beyond}, namely Penguins, which requires understanding of tabular data; Colored Object, which involves analyzing the relative positions, absolute positions, and colors of various objects; and Date, which involves understanding and calculating dates. For these three datasets, we used complete sets.

\paragraph{Code Generation}
For code generation, we selected MBPP \citep{austin2021program} and HumanEval \citep{chen2021evaluating} to construct our dataset. For MBPP, we randomly sampled 300 questions, while for HumanEval, we used its complete set.

\paragraph{Algorithmic Tasks}
We utilized Object Counting and Repeat Copy, sourced from BIG-Bench \cite{srivastava2023beyond} to evaluate the model's ability to manage details. Object Counting involves enumerating items presented within questions, while Repeat Copy requires the generation of word sequences based on instructions. The complete sets of these tasks are employed to form the detail-oriented algorithmic task of \model{}.
By utilizing the aforementioned datasets, we ensure that the questions in \model{} cover a diverse range of examination angles. The five types of tasks correspond to a variety of knowledge domains, with the detail-focused algorithmic tasks, the detail and logic-encompassing mathematical reasoning, and the logic-focused code generation corresponding to different styles of reasoning processes. Specific data statistics can be seen in Appendix \ref{appx:statistics}.

\subsection{Response Collection}
Following the collection of benchmark questions, we employ various LLMs, including GPT-3.5, GPT-4, LLaMa2 (7B, 13B, and 70B variants), and vicuna (7B, 13B, and 33B variants) to generate response for each question, using greedy decoding. The details of the prompts used are available in Appendix~\ref{appx:prompt}.

Next, we filter out the responses that did not provide valid reasoning, with a specific example provided in Table \ref{tab:invalid_response}. We then apply a random sampling strategy to maintain a consistent number of model-generated responses across each dataset.

\begin{table*}[t]
\centering
\resizebox{\linewidth}{!}{%
\begin{tabular}{lc|c|ccc|ccc}
\toprule
\multirow{2}{*}{\textbf{Model}} & \multirow{2}{*}{\textbf{Type}} & \multirow{2}{*}{\textbf{Generation}} & \multicolumn{3}{c}{\textbf{Critiquing}} & \multicolumn{3}{|c}{\textbf{Correction}} \\
                      &      &                   & ZS-AO             & ZS-CoT & FS               & ZS-CoT            & FS                 & FS (oracle)\\
\midrule
Baseline &  -   & - &  & 50.80 &  &  & 48.37  &  \\
\midrule
Phi-2 & SIFT & 45.23 & \cellcolor[HTML]{fed4c1}39.04\small{(-11.76)} & \cellcolor[HTML]{fdaa83}24.55\small{(-26.25)} & \cellcolor[HTML]{fdaa83}25.78\small{(-25.02)} & \cellcolor[HTML]{fdb897}27.69\small{(-20.68)} & \cellcolor[HTML]{fff1ea}45.39\small{(-2.98)} & \cellcolor[HTML]{cddfee}51.22\small{(+2.85)}\\
LLaMa-2-7b            & BASE & 31.66 & -  & -  & \cellcolor[HTML]{fee3d6}41.33\small{(-9.47)} & -  & \cellcolor[HTML]{fee3d6}42.27\small{(-6.10)} & \cellcolor[HTML]{cddfee}51.01\small{(+2.64)}\\
LLaMa-2-7b chat      & RLHF & 34.22 & \cellcolor[HTML]{c0d7e9}60.47\small{(+9.67)} & \cellcolor[HTML]{fff1ea}46.81\small{(-3.99)} & \cellcolor[HTML]{fee3d6}42.31\small{(-8.49)} & \cellcolor[HTML]{fdaa83}21.49\small{(-26.88)} & \cellcolor[HTML]{fee3d6}38.51\small{(-9.86)} & \cellcolor[HTML]{cddfee}51.87\small{(+3.50)}\\
Vicuna-7b            & SIFT & 31.95 & \cellcolor[HTML]{fc8d59}6.45\small{(-44.35)} & \cellcolor[HTML]{fc8d59}11.80\small{(-39.00)} & \cellcolor[HTML]{fed4c1}40.56\small{(-10.24)} & \cellcolor[HTML]{fec6ac}32.73\small{(-15.64)} & \cellcolor[HTML]{fee3d6}41.31\small{(-7.06)} & \cellcolor[HTML]{cddfee}51.56\small{(+3.19)}\\
Mistral-7b            & BASE & 47.37 & -  & -  & \cellcolor[HTML]{cddfee}55.70\small{(+4.90)} & -  & \cellcolor[HTML]{fee3d6}42.61\small{(-5.76)} & \cellcolor[HTML]{cddfee}53.23\small{(+4.86)}\\
LLaMa-2-13b            & BASE & 39.37 & -  & -  & \cellcolor[HTML]{fec6ac}32.47\small{(-18.33)} & -  & \cellcolor[HTML]{fff1ea}45.78\small{(-2.59)} & \cellcolor[HTML]{cddfee}50.88\small{(+2.51)}\\
LLaMa-2-13b chat      & RLHF & 41.67 & \cellcolor[HTML]{c0d7e9}58.41\small{(+7.61)} & \cellcolor[HTML]{fee3d6}42.87\small{(-7.93)} & \cellcolor[HTML]{fff1ea}47.79\small{(-3.01)} & \cellcolor[HTML]{fec6ac}28.89\small{(-19.48)} & \cellcolor[HTML]{fee3d6}41.67\small{(-6.70)} & \cellcolor[HTML]{cddfee}52.34\small{(+3.97)}\\
Vicuna-13b            & SIFT & 39.58 & \cellcolor[HTML]{fee3d6}40.99\small{(-9.81)} & \cellcolor[HTML]{fc8d59}11.84\small{(-38.96)} & \cellcolor[HTML]{fff1ea}46.05\small{(-4.75)} & \cellcolor[HTML]{fec6ac}30.77\small{(-17.60)} & \cellcolor[HTML]{fee3d6}42.72\small{(-5.65)} & \cellcolor[HTML]{cddfee}51.82\small{(+3.45)}\\
Vicuna-33b            & SIFT & 42.27 & \cellcolor[HTML]{fdaa83}23.96\small{(-26.84)} & \cellcolor[HTML]{fee3d6}45.64\small{(-5.16)} & \cellcolor[HTML]{cddfee}51.83\small{(+1.03)} & \cellcolor[HTML]{fee3d6}39.27\small{(-9.10)} & \cellcolor[HTML]{fee3d6}42.61\small{(-5.76)} & \cellcolor[HTML]{cddfee}52.34\small{(+3.97)}\\
LLaMa-2-70b            & BASE & 55.53 & -  & -  & \cellcolor[HTML]{cddfee}52.48\small{(+1.68)} & -  & \cellcolor[HTML]{fff1ea}46.93\small{(-1.44)} & \cellcolor[HTML]{c0d7e9}55.35\small{(+6.98)}\\
LLaMa-2-70b chat      & RLHF & 51.53 & \cellcolor[HTML]{a5c7de}67.64\small{(+16.84)} & \cellcolor[HTML]{cddfee}53.20\small{(+2.40)} & \cellcolor[HTML]{c0d7e9}59.92\small{(+9.12)} & \cellcolor[HTML]{fec6ac}30.51\small{(-17.86)} & \cellcolor[HTML]{fff1ea}44.84\small{(-3.53)} & \cellcolor[HTML]{c0d7e9}55.66\small{(+7.29)}\\
Mixtral-8×7b          & BASE & 58.43 & -  & -  & \cellcolor[HTML]{b3cfe3}\underline{63.98}\small{(+13.18)} & -  & \cellcolor[HTML]{cddfee}49.78\small{(+1.41)} & \cellcolor[HTML]{c0d7e9}56.16\small{(+7.79)}\\
Mixtral-8×7b inst & SIFT & 60.03 & \cellcolor[HTML]{fec6ac}33.36\small{(-17.44)} & \cellcolor[HTML]{fee3d6}43.34\small{(-7.46)} & \cellcolor[HTML]{cddfee}53.67\small{(+2.87)} & \cellcolor[HTML]{fee3d6}41.91\small{(-6.46)} & \cellcolor[HTML]{cddfee}\underline{51.32}\small{(+2.95)} & \cellcolor[HTML]{c0d7e9}56.44\small{(+8.07)}\\
GPT-3.5               & RLHF & 62.72 & \cellcolor[HTML]{a5c7de}\underline{69.94}\small{(+19.14)} & \cellcolor[HTML]{cddfee}51.44\small{(+0.64)} & \cellcolor[HTML]{c0d7e9}59.88\small{(+9.08)} & \cellcolor[HTML]{fff1ea}\underline{44.71}\small{(-3.66)} & \cellcolor[HTML]{cddfee}51.24\small{(+2.87)} & \cellcolor[HTML]{b3cfe3}\underline{61.22}\small{(+12.85)}\\
GPT-4                 & RLHF & 74.33 & \cellcolor[HTML]{7eafce}\textbf{81.62}\small{(+30.82)} & \cellcolor[HTML]{8bb7d3}\textbf{78.75}\small{(+27.95)} & \cellcolor[HTML]{7eafce}\textbf{86.04}\small{(+35.24)} & \cellcolor[HTML]{c0d7e9}\textbf{56.65}\small{(+8.28)} & \cellcolor[HTML]{98bfd9}\textbf{69.96}\small{(+21.59)} & \cellcolor[HTML]{8bb7d3}\textbf{74.80}\small{(+26.43)}\\
\bottomrule
Average   &  -    & 47.73 & \cellcolor[HTML]{fff1ea}48.19\small{(-2.61)} & \cellcolor[HTML]{fee3d6}41.02\small{(-9.78)} & \cellcolor[HTML]{fff1ea}50.65\small{(-0.15)} & \cellcolor[HTML]{fed4c1}35.46\small{(-12.91)} & \cellcolor[HTML]{fff1ea}46.46\small{(-1.91)} & \cellcolor[HTML]{c0d7e9}55.06\small{(+6.69)}\\
\midrule
Auto-J-13b                & CT   & -     & -     & \cellcolor[HTML]{b3cfe3}\underline{65.29}\small{(+14.49)} & -     & -     & -      & -    \\
UltraCM-13b               & CT   & -     & -     & \cellcolor[HTML]{b3cfe3} 61.11\small{(+10.31)} & -     & -     & -      & -    \\
\bottomrule
\end{tabular}
}
\caption{
Average performance on \model{}. In the table, the values in parentheses under the "Critiquing" column show comparisons to the Baseline critique score of random guessing (50.80). Similarly, in the "Correction" column, the parentheses display changes relative to the Baseline generation score from the original response (48.37) in \model{}. Blue highlights indicate improvement, while orange marks decline. Type: BASE refers to the pretrained model, SIFT means its enhancement via Supervised Instruction Finetuning, RLHF denotes further training with Reinforcement Learning from Human Feedback, and CT represents Critique Training.}
\label{tab:main_tab}
\end{table*}

\subsection{Response Annotation}
Response correctness is initially determined by rule-based matching, followed by a more detailed evaluation using GPT-4 to assess response precision. This includes flagging mathematically correct answers with incorrect reasoning and recognizing near-correct commonsense responses. Discrepancies between GPT-4 evaluations and initial annotations are resolved through manual review. During this review, we identified questions in the Date dataset that lack correct options, with a detailed example provided in Figure \ref{tab:error_in_date}. Examples of annotations are provided in Appendix \ref{appx:annotation_example}.
\subsection{Evalution}
\label{sec:evalution}
\noindent{\textbf{Evaluation Process}}
The evaluation process on \model{} is illustrated in Figure \ref{fig:flow}. First, a critique prompt is constructed using the response within \model{}, prompting the model to perform a critique. Subsequently, a correct prompt is built incorporating the critique to obtain the model's correction results.

\noindent{\textbf{Generation and Correction Metrics}}
We use the accuracy metric $S_a$ to assess models' generation and correction capability as follows:
\begin{equation}
S_a = \frac{c}{N}
\end{equation}
where $c$ is the number of correct predictions, $N$ is the total number of questions.

\noindent{\textbf{Critique Metrics}}
We assess the critique ability of LLMs by prompting them to evaluate the correctness of given responses. To address potential issues like class imbalance, biases \cite{wang2023large}, and the unreliability \cite{gou2024critic} of LLM-based evaluations, we utilize the F1 score as a more robust and reproducible metric for critiquing errors:

\begin{equation}
S_p = \frac{\sum_{i=1}^{m}q_i}{m},
\end{equation}
\begin{equation}
S_r = \frac{\sum_{i=1}^{n}q_i}{n},
\end{equation}
\begin{equation}
q_i = \begin{cases}
1, & \text{if correctly classified as wrong}, \\
0, & \text{otherwise}.
\end{cases}
\end{equation}
\begin{equation}
S_f = 2 \times \frac{S_p \times S_r}{S_p + S_r},
\end{equation}
where $S_p$ is the precision score, $S_r$ is the recall score, $m$ is the number of classified as wrong, $n$ is the number of actual wrong, and $q_i$ indicates the correct discrimination of a response as wrong.

\section{Experiments}
\subsection{Experimental Setup}
To conduct a comprehensive assessment, we have selected the following models: Phi-2, the LLaMa family, the Vicuna family, the  Mistral family, and the GPT family. We evaluate those models on \model{} under three phases:
(1) \emph{Generation}: LLMs are tasked with answering questions through the utilization of CoT \citep{wei2022chain}, during which accuracy serves as the performance indicator.
(2) \emph{Critique}: This phase evaluates provided response correctness. To fairly evaluate the critique ability of LLMs, we use a fixed set of responses from \model{} as input data, along with the F1 score outlined in Section \ref{sec:evalution} as our critique metric to avoid potential issues like class imbalance, biases, and the unreliability \citep{wang2023large} of LLM-based evaluations, aiming for a more equitable assessment. After testing three prompts from \citet{huang2023large}, we selected the most effective for zero-shot use. 
By modifying the original prompt, we experimented with two prompts: a zero-shot answer only (ZS-AO) and a zero-shot chain of thought (ZS-COT) requesting analysis. 
For pre-trained models with weaker instruction adherence (labeled BASE), we focused on few-shot (FS) setting.
Additionally, to assess critique training effects, we evaluated the 13B model Auto-J \citep{li2023generative} and UltraCM \citep{cui2023ultrafeedback} by extracting their discrimination on the response using a rule-based method.
(3) \emph{Correction}: After the critique phase, this phase refines responses by addressing identified inaccuracies in the generated critique. In this stage, accuracy is the metric used for evaluation. Besides ZS-COT and FS, we also evaluated FS (oracle), where corrections were applied only to those responses identified as incorrect in the \model{}. 

The prompts for above phases are detailed in Appendix \ref{appx:prompt}. For all experimental setups, we set the temperature to 0 during the three phases.
 \begin{figure}
    \centering
    \includegraphics[width=0.9\linewidth]{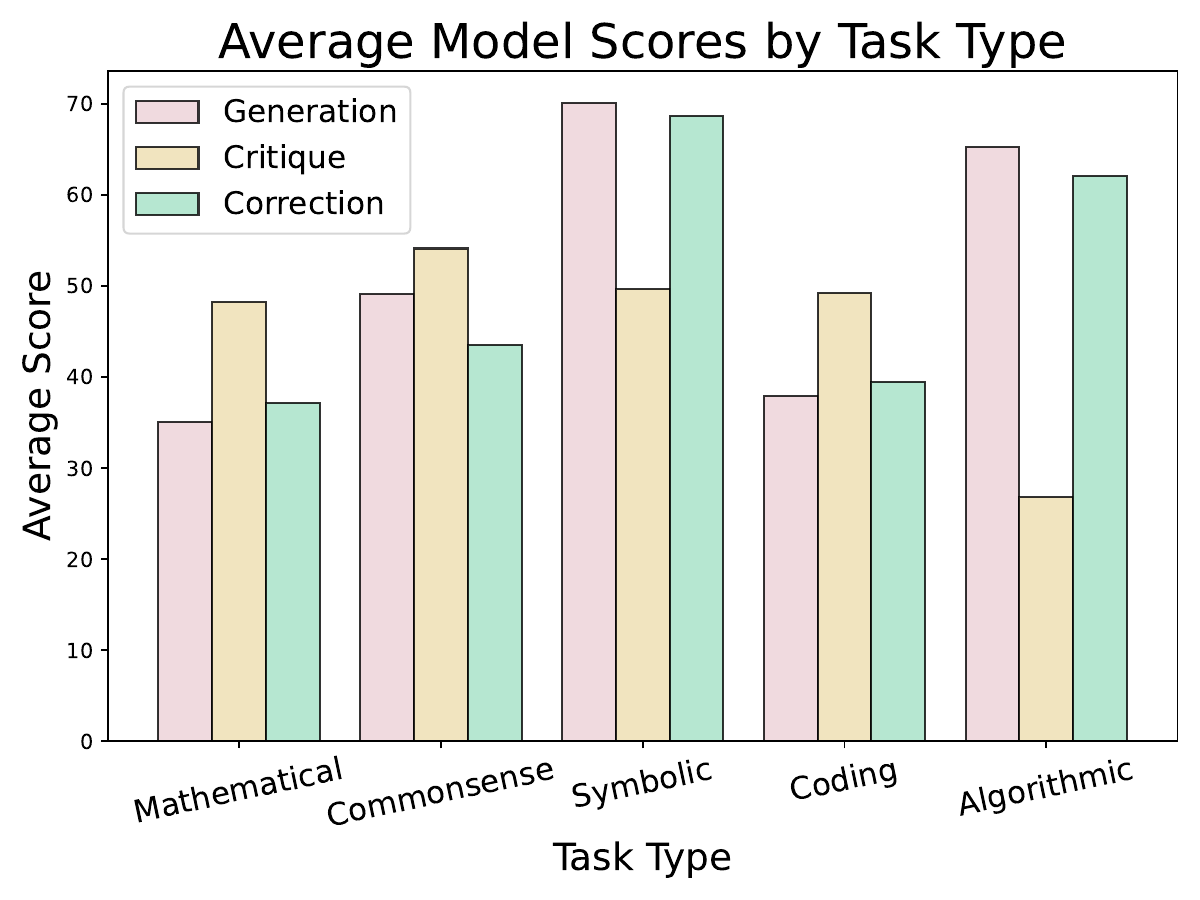}
    \caption{Average Score on different types of tasks.}
    \label{fig:ava_on_diff_type}
\end{figure}
\subsection{Results and Analysis}
Table \ref{tab:main_tab} showcases the performance of LLMs on \model{}. Specifically, we are interested in exploring the following research questions: 
\textbf{RQ1:} What factors influence the model's generation, critique, and correction?
\textbf{RQ2:} What's the interrelationship between a model's capabilities in generation, critique, and correction?
\textbf{RQ3:} How do critique-correct reasoning capabilities differ across various task types?
\textbf{RQ4:} Is the model's knowledge consistent in generation, critique, and correction?
\textbf{RQ5:} How does the inter-model critiquing patterns manifest among models of varying capability?
In the following sections, we will discuss these research questions in turn.
\begin{figure*}[t]
    \centering
    \includegraphics[width=0.9\linewidth]{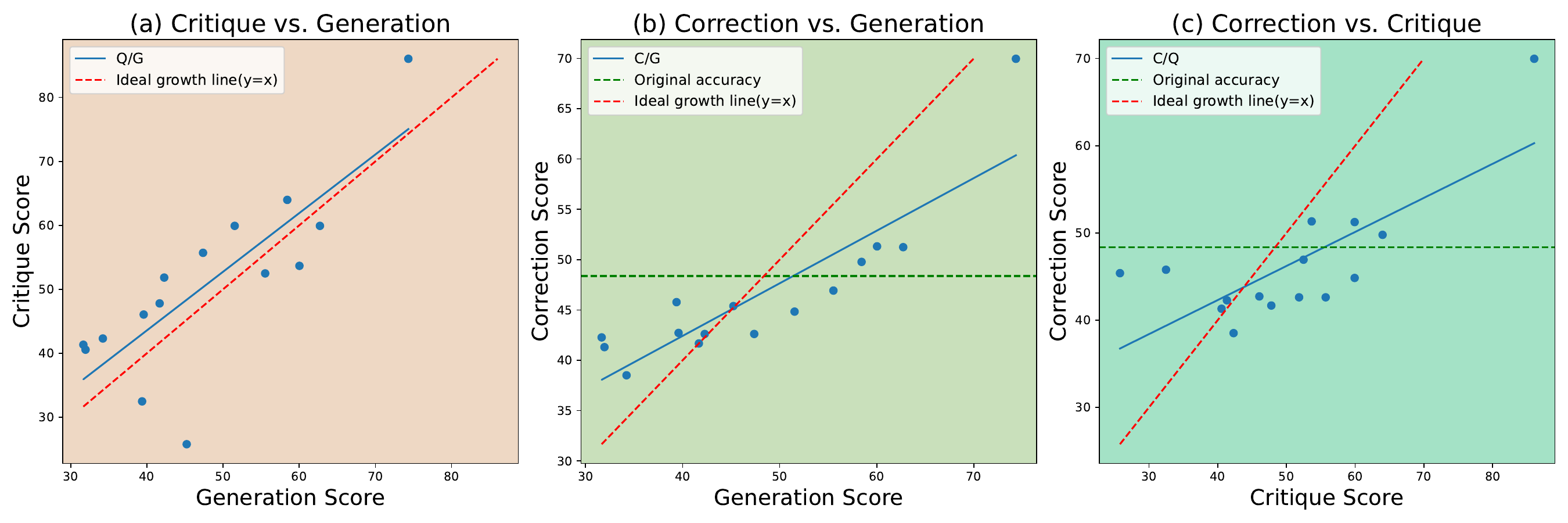}
    \caption{Interrelationship between a model’s capabilities in generation, critique, and correction. Each point on the graph represents a model, with coordinates indicating its performance in Generation (G), Critique (Q), and Correction (C). The graph features fitted lines for the scatter plots, denoted by blue lines for Q/G, C/G, and C/Q, while a red dashed line represents the ideal growth line (y=x). The green dashed line marks the original accuracy of responses from \model{}.}
    \label{fig:total_interconnectedness}
\end{figure*}
\subsubsection{RQ1: Key Factors in LLM Critical Reasoning}

\paragraph{Base Model \& Scale}
Observations reveal that Phi-2 (2.7B), despite excelling in generation tasks, exhibits weaker performance in critique tasks compared to models with similar generation performance (\emph{e.g.}, LLaMa-2-13b, Vicuna-33b). 
The Phi series \citep{gunasekar2023textbooks} is trained using high-quality data sourced from the web, along with textbooks and exercises synthesized for downstream tasks based on GPT-3.5. This training focuses primarily on enhancing capabilities in reasoning, language understanding, and knowledge  \citep{li2023textbooks}, which may contribute to its relative lack of proficiency in critique tasks.
This underscores the necessity of evaluating generation, critique, and correction collectively to achieve a comprehensive assessment of a model's mastery of knowledge. Additionally, Mistral-7b stands out as the top performer among models of similar size, even outperforming Vicuna-33b. However, in Figure \ref{fig:rader}, GPT-4 consistently maintains a significant lead in GQC of all types of tasks. Despite this, other models like LLaMA-70b and Mixtral-8×7b demonstrate competitiveness against GPT-3.5. 

Furthermore, it is observed that models with more than 13 billion parameters exhibit certain critique capabilities (surpassing the baseline of random guessing). Meanwhile, only models of the Mixtral-8×7b and above are capable of achieving effective correction (exceeding the baseline generation score).

\paragraph{Training Strategy}
By comparing the results of different models under LLaMa family, it is observed that the alignment tax has limited the RLHF's generation performance. However, for critique and correction, RLHF consistently outperforms BASE, suggesting that RLHF might suppress the expression of knowledge in generation. Simultaneously, in critique, CT demonstrates results surpassing those of GPT-3.5 with a smaller parameter size (13B), proving the effectiveness of CT.

\paragraph{Prompt Strategy}
The critique results from zero-shot settings show sensitivity to prompts. For instance, Vicuna-13b's ZS-AO flagged 22.04\% of responses as incorrect, compared to only 4.8\% in ZS-CoT, against an actual error rate of 51.63\%. This inconsistency likely stems from the model's insufficient training on critique tasks, making it struggle without clear examples. Meanwhile, in correction, few-shot also significantly outperforms zero-shot. 
Therefore, FS is always the better choice in both critique and correction. For subsequent analysis, we will primarily focus on FS results. 

\paragraph{Orcale Feedback}
We also explored an oracle setting for corrections by modifying only incorrect responses in \model{}. Results from FS (oracle) outperformed those without an oracle, showing that reliable external feedback can significantly enhance correction efficiency. However, for more advanced models (from LLaMa-2-70b to GPT-4), corrections in the oracle setting still fell short of direct generation, indicating they are still influenced by incorrect responses from other models.

\subsubsection{RQ2: Correlations of GQC Abilities}
\emph{The capabilities of generation, critique, and correction exhibit a positive correlation.}
Figures \ref{fig:total_interconnectedness} illustrates the interconnectedness among three capabilities. 
It is observed that there is a positive linear relationship between generating and judging. The improvement rates of generation and critique are nearly identical, even though the model primarily focuses on learning tasks related to generation during training.
However, the linear correlation between the generation and correction capacities is not prominent. Weaker models tend to exhibit diminished correctness upon correction, as compared to the initial benchmark responses. 
This observation suggests that a model’s limited capability to generate precise answers may impact its ability to correct those responses. Similarly, while the relationship between critique and correction shows that the model can discriminate between correct and incorrect responses, it appears that it may not always be capable of rectifying them.
\begin{figure*}[t]
    \centering
    \includegraphics[width=0.9\linewidth]{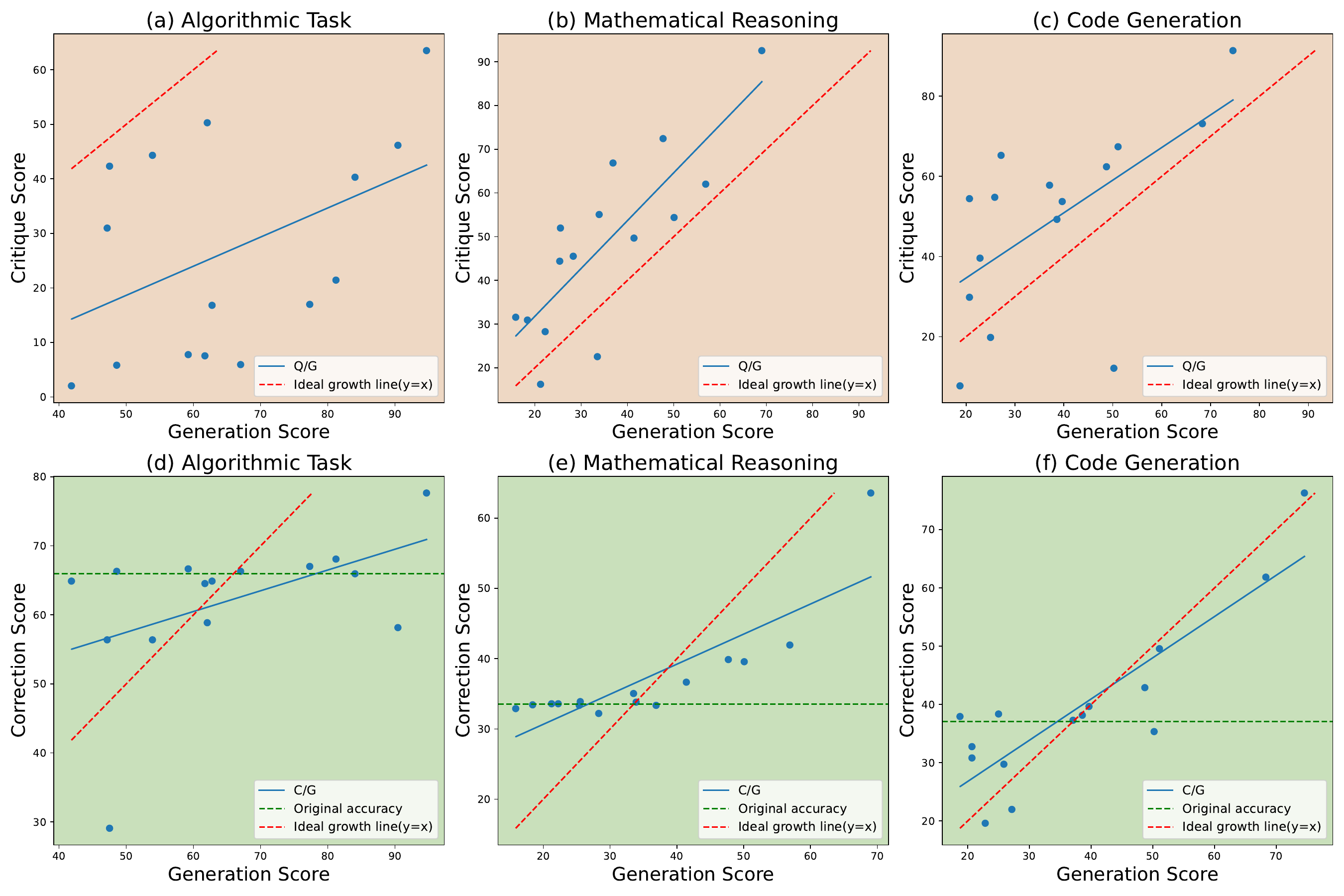}
    \caption{Critique/Generation (Q/G) and Correction/Generation (C/G) line in Algorithmic Task, Mathematical Reasoning, and Code Generation. The green dashed line marks the original accuracy of
responses from CRITICBENCH}
    \label{fig:type_interconnectedness}
\end{figure*}

\subsubsection{RQ3: Impact of Task Type}
\emph{The model's critique and correction capability depends on whether the task focuses on details or logic.}
In Figure \ref{fig:type_interconnectedness}, we illustrate the variability in critique and correction capabilities across various task types. 
Figure \ref{fig:type_interconnectedness} (a), (b), and (c) demonstrate the varying relationships between Q and G across different types of tasks. Specifically, the models exhibit weaker critique performance in detail-oriented algorithmic tasks compared to their generation abilities, indicated by dots below the ideal growth line y=x. In contrast, for mathematical reasoning and code generation tasks, their critique capabilities surpass generation capabilities. The ability to correct errors in algorithmic tasks is also limited, even when the model's generation performance significantly surpasses that of the original responses. As demonstrated in Figure \ref{fig:type_interconnectedness} (e), improvement in correction effectiveness for mathematical tasks, which require detailed and logical reasoning, is only observed when the model's direct generation accuracy exceeds 41\%. This suggests that in mathematical tasks, models are also susceptible to being misled by incorrect responses, which may explain the lack of improvement of math in the self-refine \citep{madaan2023self}. Interestingly, For the logic-focused code generation tasks, improvements are realized as long as the model's generation performance surpasses that of the original responses' performance. This result indicates that models, when performing critique and correction, are easily disrupted by incorrect answers in tasks that focus on details, but not in those that emphasize logic. For examples of tasks focusing on details and logic, see Appendix \ref{appx:Examples of Detail-Oriented and Logic-Oriented tasks}.

Additionally, Figure \ref{fig:ava_on_diff_type} displays the model's average GQC score on different types of tasks. It can be observed that, similar to algorithmic tasks, the model struggles to effectively critique in detail-oriented symbolic reasoning, where its performance is significantly lower than that in generation and correction. This further demonstrates the model's lack of critique ability on such tasks. For more detailed results, please refer to Appendix \ref{appx:Detailed Results on Different Tasks}.
\begin{figure}[ht]
    \centering
    \includegraphics[width=0.9\linewidth]{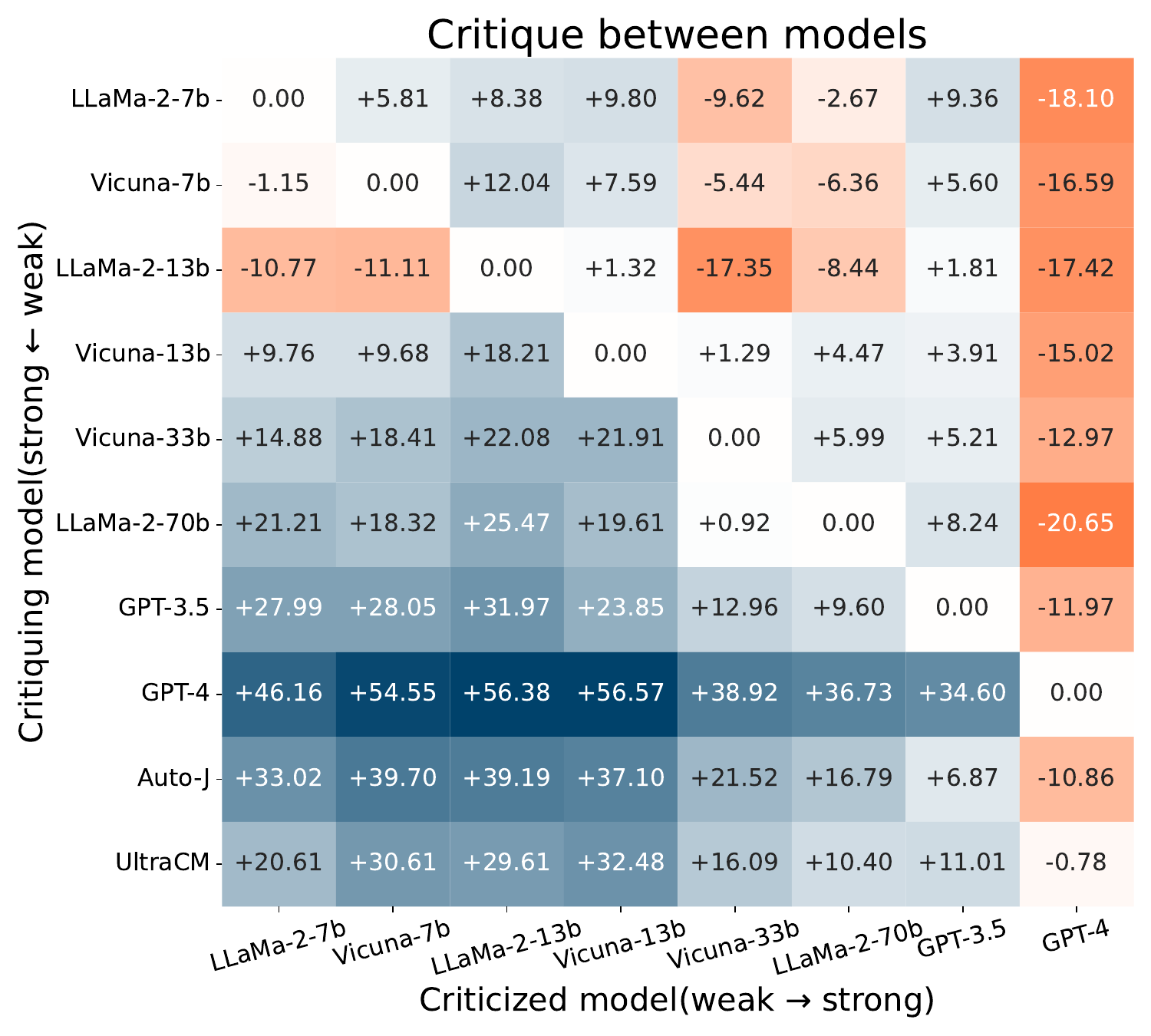}
    \setlength{\belowcaptionskip}{-10pt}
    \caption{The results of model critiques are depicted in the graph, where the self-critique scores of LLMs are set to 0 for comparison. Models are arranged from weakest to strongest in order of generation accuracy.}
    \label{fig:weak_to_strong}
\end{figure}

\subsubsection{RQ4: Consistency of GQC Knowledge}
\emph{GQC knowledge inconsistencies persist across all models.}
The relationship between human abilities of generation, critique, and correction suggests that generation falls under critique, with correction closely linked to generation \citep{west2024the}. This implies humans can identify errors without necessarily knowing the correct answer, whereas generating the right answer also implies the ability to critique and correct it. However, the knowledge acquired by LLMs is not entirely consistent across generation, critique, and correction tasks.

Figure \ref{fig: g_d_c_distribution} delineates the overlap and distinctiveness of GQC, highlighting their inconsistencies of knowledge. As a model's parameter size increases, its knowledge coherence across GQC, and instances of complete task failure (where generation, critique, and correction all fail) decrease. Notably, it can be noted that questions correctly critiqued alone always occupy a significant portion, indicating that the model possesses a considerable amount of knowledge that is not expressed through generation or correction. Furthermore, the green segment in Figure \ref{fig: g_d_c_distribution} demonstrates a counterintuitive phenomenon: while the model successfully provides correct responses in generation and correction, it fails to accurately judge the correctness of the given responses in critique.

\subsubsection{RQ5: Patterns of Inter-Model Critique}

Figure \ref{fig:weak_to_strong} presents a visualization illustrating the inter-model critiquing result.
Overall, it can be observed that strong models consistently have a superior ability to critique than weak models, and the responses of weak models are more easily critiqued accurately, possibly because the errors made by weak models are more evident. Interestingly, certain weaker models approach or even exceed the self-critique scores of stronger models, suggesting that their critique capacity against stronger models might surpass the latter's self-critique. Post-critique training, models such as Auto-J and UltraCM show enhanced ability to assess response correctness across different models, with UltraCM's performance nearing its self-critique level against GPT-4, underscoring the value of critique training.

\section{Conclusion}

In conclusion, our investigation through \model{} has illuminated the capacities and limitations of LLMs in GQC reasoning. 
Our focused exploration using \model{} on the relationship among models in GQC revealed a linear correlation and subtle inconsistencies between GQC, while our analysis across different task types found that models perform better in Q and C for tasks focused on logic compared to those requiring attention to detail. Additionally, by examining the outcomes of models critiquing each other, we discovered that weaker models could sometimes correct the outputs of stronger models more effectively than those models could self-correct. These findings underscore the effectiveness of \model{} in evaluating and analyzing the GQC capabilities of LLMs.

\section*{Limitations}
Measuring the ability of model critique effectively has always been a challenge. In this paper, we use discrimination results as a valid indicator to measure its critique ability, mainly for the following reasons: (1) Fine-grained indicators that provide scores based on evaluation principles are only suitable for specific tasks and lack generality. Moreover, different tasks have different focuses, and evaluation principles valued by humans may dynamically change. It is somewhat idealistic to exhaustively list all evaluation principles once and for all. (2) Scores based on evaluation principles rely on human annotations or results from GPT-4 for validation. However, reliable human annotations incur high costs, and GPT-4 may contain errors and biases. Using GPT-4 results also makes it impossible to evaluate its critique ability. Additionally, in reasoning tasks, the most important aspect of critique and correction is to judge whether there are errors in the reasoning process and its results. Therefore, considering the above considerations, we use a binary metric to measure the results.

Future work should address these challenges by exploring alternative evaluation methodologies that mitigate reliance on costly human annotations. Additionally, there is a need to develop more nuanced critique metrics that can effectively capture the diverse aspects of model performance across various tasks and evaluation scenarios.

\section*{Ethics Statement}
We constructed the model based on existing public datasets and models, as detailed in Section \ref{sec:criticbench}, and annotated the results using GPT-4 and human evaluation. We acknowledge that, despite employing rule-based filtering, GPT-4 review, and human review, unpredictable errors may still exist in the responses generated by different models.

Additionally, when using the critique ability of Large Language Models (LLMs), it's also important to be aware of the risks involved, such as potential biases. When the GQC capability of an LLM is inconsistent, and its critique ability surpasses the other two aspects, it is necessary to carefully discern whether its discriminate results contain harmful biases.

\section*{Acknowledgements}
This work was partly supported by the National Key Research and Development Program of China (No. 2020YFB1708200) , the "Graph Neural Network Project" of Ping An Technology (Shenzhen) Co., Ltd. and the Shenzhen Science and Technology Program (JSGG20220831110203007).

\bibliography{anthology,custom}

\appendix
\onecolumn
\newpage
\addtocontents{toc}{\protect\setcounter{tocdepth}{3}}

\renewcommand{\contentsname}{Contents of the Appendix}  
\tableofcontents %

\newpage
\section{Statistics of \model{}}
\label{appx:statistics}
\begin{table*}[htbp]  
\centering  
\begin{tabular}{lccc}  
\toprule  
\textbf{Type}                & \textbf{Dataset}                  & \textbf{Samples} & \textbf{License}\\  
\midrule  
\multirow{4}{*}{Math}        & GSM8K          & 250                        & MIT License           \\  
                             & MATH           & 500                        & MIT License           \\  
                             & AQuA           & 254                        & Apache License 2.0    \\  
                             & TabMWP         & 300                        & CC BY-NC-SA 4.0       \\  
\midrule  
\multirow{4}{*}{Commonsense} & CSQA           & 300                        & MIT License           \\  
                             & AmbigNQ        & 300                        & CC BY-SA 3.0          \\  
                             & StrategyQA     & 229                        & MIT License           \\  
                             & HotpotQA       & 300                        & CC BY-SA 4.0          \\  
\midrule  
\multirow{3}{*}{Symbolic}    & Penguins       & 146                        & Apache License 2.0    \\  
                             & Colored Object & 250                        & Apache License 2.0    \\  
                             & Date           & 250                        & Apache License 2.0    \\  
\midrule  
\multirow{2}{*}{Coding}      & MBPP           & 300                        & CC BY 4.0             \\  
                             & HumanEval      & 164                        & MIT License           \\  
\midrule  
\multirow{2}{*}{Algorithmic} & Object Counting & 250                       & Apache License 2.0    \\  
                             & Repeat Copy    & 32                         & Apache License 2.0    \\  
\midrule  
All                          & -              & 3825                       & -                     \\  
\bottomrule  
\end{tabular}  
\caption{Statistics of the datasets used in \model{}.}  
\label{tab:dataset_detail}  
\end{table*}  

\section{Annotation Example}
\label{appx:annotation_example}
\begin{table*}[htbp]
\centering
\resizebox{\linewidth}{!}{%
\begin{tabular}{p{\textwidth}}
\toprule
\textbf{Question:} Machine A puts out a yo-yo every 6 minutes. Machine B puts out a yo-yo every 9 minutes. After how many minutes will they have produced 10 yo-yos?\\
Answer Choices: A)24 minutes B)32 minutes C)36 minutes D)64 minutes E)72 minutes \\ 
\midrule
\textbf{Solution:} Machine A puts out a yo-yo every 6 minutes. Machine B puts out a yo-yo every 9 minutes. So Machine A will produce 10 yo-yos in 60 minutes. Machine B will produce 10 yo-yos in 90 minutes. So Machine A will produce 10 yo-yos in 60 minutes and Machine B will produce 10 yo-yos in 90 minutes. So after 120 minutes, they will have produced 10 yo-yos. The answer is C) 120.\\
\midrule
\textbf{Answer:} C	\\
\midrule
\textbf{Rule-based answer matching:} Correct	\\
\midrule
\textbf{GPT-4 evaluation:} Wrong	\\
\midrule
\textbf{Manual review:} Wrong\\
\midrule
\textbf{Final label:} Wrong\\
\bottomrule
\end{tabular}
}
\caption{Inconsistencies between rule-based matching and GPT-4 evaluations (AQuA).}
\end{table*}

\begin{table*}
\centering
\begin{adjustbox}{width=\textwidth}

\begin{tabular}{p{\linewidth}}
\toprule
\textbf{Question:} 
\begin{lstlisting}[style=Python]
def words_in_sentence(sentence):
    """
    You are given a string representing a sentence,
    the sentence contains some words separated by a space,
    and you have to return a string that contains the words from the original sentence,
    whose lengths are prime numbers,
    the order of the words in the new string should be the same as the original one.

    Example 1:
        Input: sentence = "This is a test"
        Output: "is"

    Example 2:
        Input: sentence = "lets go for swimming"
        Output: "go for"

    Constraints:
        * 1 <= len(sentence) <= 100
        * sentence contains only letters
    """ 
\end{lstlisting} \\
\midrule
\textbf{Solution:} 
\begin{lstlisting}[style=Python]
def words_in_sentence(sentence):
    words = sentence.split()
    prime_words = [word for word in words if len(word) in [2, 3, 5, 7]]
    return ' '.join(prime_words)
\end{lstlisting} \\
\midrule
\textbf{Check function:}
\begin{lstlisting}[style=Python]
def check(candidate):
    # Check some simple cases
    assert candidate("This is a test") == "is"
    assert candidate("lets go for swimming") == "go for"
    assert candidate("there is no place available here") == "there is no place"
    assert candidate("Hi I am Hussein") == "Hi am Hussein"
    assert candidate("go for it") == "go for it"
    # Check some edge cases that are easy to work out by hand.
    assert candidate("here") == ""
    assert candidate("here is") == "is"
check(words_in_sentence)
\end{lstlisting} \\
\midrule
\textbf{Check result:} Correct\\
\midrule
\textbf{GPT-4 evaluation:} Wrong	\\
\midrule
\textbf{Manual review:} Wrong\\
\midrule
\textbf{Final label:} Wrong\\
\bottomrule
\end{tabular}
\end{adjustbox}
\caption{Inconsistencies between rule-based matching and GPT-4 evaluations (HumanEval).}
\end{table*}

\begin{table*}
\centering
\resizebox{\linewidth}{!}{%
\begin{tabular}{p{\textwidth}}
\toprule
\textbf{Question:} Lake Wales Ridge National Wildlife Refuge is on the US route that runs between a city in what Florida county?\\
\midrule
\textbf{Solution:} Lake Wales Ridge National Wildlife Refuge is on the US route that runs between a city in Polk County. The answer is Polk County.\\
\midrule
\textbf{Answer:} polk\\
\midrule
\textbf{Rule-based answer matching:} Wrong\\
\midrule
\textbf{GPT-4 evaluation:} Correct	\\
\midrule
\textbf{Manual review:} Correct\\
\midrule
\textbf{Final label:} Correct\\
\bottomrule
\end{tabular}
}
\caption{Inconsistencies between rule-based matching and GPT-4 evaluations (HotpotQA).}
\end{table*}

\newpage

\section{Example of Invalid Response}

\label{appx:invalid response}
\begin{table*}[htbp]
\centering
\resizebox{\linewidth}{!}{%
\begin{tabular}{p{\textwidth}}
\toprule
\textbf{Question:} The number of students in a school hall was 1000. 
The hall had 3 entrances A, B, and C which also served as the exits. 
If after a meeting 30\% of the students went out of the building through exit A, 
3/5 of the remaining went out through exit B, and the rest went out through exit C, 
calculate the number of students who went out through exit C.\\
Answer Choices: A)24 minutes B)32 minutes C)36 minutes D)64 minutes E)72 minutes \\ 
\midrule
\textbf{Solution:} 1000 students in the hall. 30\% of the students went out through exit A. 
3/5 of the remaining went out through exit B. 
So 1000 - 30\% - 3/5 = 1000 - 30\% - 3/5 = 1000 - 30\% - 3/5 = 1000 - 30\% - 3/5 = 1000 - 30\% - 3/5 = 1000 - 30\% - 3/5 = 1000 - 30\% - 3/5 = 1000 - 30\% - 3/5 = 1000 - 30\% -\\
\bottomrule
\end{tabular}
}
\caption{Example of excluded invalid responses.}
\label{tab:invalid_response}
\end{table*}

\section{Manual Quality Check}
\label{appx:manual_quality_check}
We have conducted detailed manual checks to ensure the label correctness of the \model{}.  
\eg during our examination, we discovered that all questions in the dataset pertaining to the event "Jane and John married on Jan 2, 1958," have issues with their options. The specific errors are detailed in Table \ref{tab:error_in_date}.

\begin{table*}[htbp]
\centering
\resizebox{\linewidth}{!}{
\begin{tabular}{p{\textwidth}}
\toprule
\textbf{Question: }Jane and John married on Jan 2, 1958. It is their 5-year anniversary today. What is the date today in MM/DD/YYYY? \\
Options: \\
(A) 01/02/1894 (B) 01/02/1961 (C) 12/23/1960 (D) 01/02/1960 (E) 01/03/1961 (F) 01/02/1990 \\
\midrule
\textbf{Provided Answer:} B \\
\midrule
\textbf{Explanation:} If Jane and John married on Jan 2, 1958, the date of Jane and John's 5-year anniversary should be Jan 2, 1963. \\
\bottomrule
\end{tabular}
}
\caption{Incorrect Question in Date (no correct options provided).}
\label{tab:error_in_date}
\end{table*}

\newpage
\section{Examples of Detail-Oriented and Logic-Oriented tasks}
\label{appx:Examples of Detail-Oriented and Logic-Oriented tasks}
Detail-oriented tasks require meticulous examination of questions or strict adherence to specific instructions. In contrast, logic-oriented tasks necessitate logical reasoning.

Algorithmic Tasks typically demand a focus on details—careful attention to the question's conditions to ensure the response aligns with those specifics. Mathematical Reasoning requires a balance of both—attention to the problem's constraints and logical reasoning to ensure the calculation process is accurate. Meanwhile, Code Generation primarily hinges on logical structuring, with the correctness of outcomes mainly influenced by the logic embedded in the code, thus categorized under logic-oriented tasks. Specific examples can be found in Tables \ref{tab:detail-oriented tasks} and \ref{tab:logic-oriented tasks}.

\begin{table*}[!htbp]
\centering
\resizebox{\linewidth}{!}{%
\begin{tabular}{p{\textwidth}}
\toprule
\textbf{Question:} Repeat the phrase all cars eat gas four times. 
On the odd times, drop words that start with vowels.\\
\midrule
\textbf{Answer:} cars gas all cars eat gas cars gas all cars eat gas\\
\midrule
\textbf{Type:} Algorithmic Task\\
\bottomrule
\end{tabular}
}
\caption{Example of detail-oriented tasks.}
\label{tab:detail-oriented tasks}
\end{table*}

\begin{table*}[!htbp]
\centering
\begin{adjustbox}{width=\textwidth}
\begin{tabular}{p{\textwidth}}
\toprule
\textbf{Question:} Write a function to find the volume of a cuboid.\\
\midrule
\textbf{Answer:} \\
\begin{lstlisting}[style=Python]
def volume_cuboid(length, width, height):
    return length * width * height
\end{lstlisting} \\

\midrule
\textbf{Type:} Code Generation\\
\bottomrule
\end{tabular}
\end{adjustbox}
\caption{Example of logic-oriented tasks.}
\label{tab:logic-oriented tasks}
\end{table*}

\newpage
\section{Detailed Results on Different Tasks}
\label{appx:Detailed Results on Different Tasks}
\begin{table}[!htbp]
\centering
\begin{tabular}{lccc}
\toprule
            \textbf{Model} &  \textbf{Generation} &  \textbf{Critiquing} &  \textbf{Correction} \\
\midrule
    Baseline &       - &       57.08 &       33.51 \\
\midrule
Phi-2 & 33.51& 22.55& 35.05\\
LLaMa-2-7b & 15.87& 31.57& 32.90\\
LLaMa-2-70b chat & 21.24& 16.23& 33.59\\
Vicuna-7b & 18.40& 30.93& 33.44\\
Mistral-7b & 33.90& 55.07& 33.82\\
LLaMa-2-13b & 22.24& 28.27& 33.59\\
LLaMa-2-70b chat & 28.30& 45.52& 32.21\\
Vicuna-13b & 25.38& 44.39& 33.36\\
Vicuna-33b & 25.54& 51.97& 33.90\\
LLaMa-2-70b & 41.41& 49.67& 36.66\\
LLaMa-2-70b chat & 36.89& 66.85& 33.36\\
Mixtral-8×7b & 47.70& \underline{72.43}& 39.88\\
Mixtral-8×7b inst & 50.08& 54.38& 39.57\\
GPT-3.5 & \underline{56.90}& 62.01& \underline{41.95}\\
GPT-4 & \textbf{69.02}& \textbf{92.55}& \textbf{63.57}\\
\midrule
Average score & 35.09& 48.29& 37.12\\
\bottomrule
\end{tabular}
\caption{Mathematical Reasoning Task Performance}
\label{tab:math_reasoning}
\end{table}
\begin{table}[!htbp]
\centering
\begin{tabular}{lccc}
\toprule
            \textbf{Model} &  \textbf{Generation} &  \textbf{Critiquing} &  \textbf{Correction} \\
\midrule
    Baseline &       - &       46.82 &       55.98 \\
\midrule
Phi-2 & 37.56& 37.81& 42.96\\
LLaMa-2-7b & 41.45& 56.99& 37.47\\
LLaMa-2-70b chat & 43.05& 60.93& 34.01\\
Vicuna-7b & 37.38& 54.64& 34.01\\
Mistral-7b & 49.25& \underline{62.96}& 36.76\\
LLaMa-2-13b & 49.51& 48.42& 45.26\\
LLaMa-2-70b chat & 48.36& 51.95& 39.15\\
Vicuna-13b & 42.69& 51.05& 43.49\\
Vicuna-33b & 50.58& 47.39& 43.67\\
LLaMa-2-70b & \underline{58.72}& 57.03& 46.94\\
LLaMa-2-70b chat & 56.24& 55.12& 46.24\\
Mixtral-8×7b & 55.36& 59.07& 47.92\\
Mixtral-8×7b inst & 56.07& 46.77& \underline{50.75}\\
GPT-3.5 & 46.41& 50.22& 44.11\\
GPT-4 &\textbf{64.39}& \textbf{71.56}& \textbf{59.96}\\
\midrule
Average score & 49.13& 54.13& 43.51\\
\bottomrule
\end{tabular}
\caption{Commonsense Reasoning Task Performance}
\label{tab:commonsense_reasoning}
\end{table}
\begin{table}[!htbp]
\centering
\begin{tabular}{lccc}
\toprule
            \textbf{Model} &  \textbf{Generation} &  \textbf{Critiquing} &  \textbf{Correction} \\
\midrule
    Baseline &       - &       40.84 &       65.48 \\
\midrule
Phi-2 & 69.20& 26.80& 68.58\\
LLaMa-2-7b & 48.92& 45.12& 66.56\\
LLaMa-2-70b chat & 48.92& 32.62& 65.94\\
Vicuna-7b & 53.56& 48.75& 65.79\\
Mistral-7b & 70.90& 51.77& 64.09\\
LLaMa-2-13b & 62.54& 24.83& 67.65\\
LLaMa-2-70b chat & 63.00& 31.48& 67.34\\
Vicuna-13b & 65.17& 55.66& 67.34\\
Vicuna-33b & 65.02& 53.17& 62.69\\
LLaMa-2-70b & 80.34& 56.77& 64.09\\
LLaMa-2-70b chat & 78.64& 53.49& 64.86\\
Mixtral-8×7b & 82.51& 56.87& 69.97\\
Mixtral-8×7b inst & 82.97& 52.66& 70.90\\
GPT-3.5 & \underline{86.84}& \underline{64.49}& \underline{71.83}\\
GPT-4 & \textbf{93.34}& \textbf{90.75}& \textbf{92.41}\\
\midrule
Average score & 70.12& 49.68& 68.67\\
\bottomrule
\end{tabular}
\caption{Symbolic Reasoning Task Performance}
\label{tab:symbolic_reasoning}
\end{table}
\begin{table}[!htbp]
\centering
\begin{tabular}{lccc}
\toprule
            \textbf{Model} &  \textbf{Generation} &  \textbf{Critiquing} &  \textbf{Correction} \\
\midrule
    Baseline &       - &       55.73 &       37.07 \\
\midrule
Phi-2 & 50.22& 12.14& 35.34\\
LLaMa-2-7b & 18.75& 7.74& 37.93\\
LLaMa-2-70b chat & 20.69& 54.43& 30.82\\
Vicuna-7b & 20.69& 29.83& 32.76\\
Mistral-7b & 38.58& 49.27& 38.15\\
LLaMa-2-13b & 25.00& 19.82& 38.36\\
LLaMa-2-70b chat & 25.86& 54.77& 29.74\\
Vicuna-13b & 22.84& 39.59& 19.61\\
Vicuna-33b & 27.16& 65.23& 21.98\\
LLaMa-2-70b & 39.66& 53.72& 39.66\\
LLaMa-2-70b chat & 37.07& 57.79& 37.28\\
Mixtral-8×7b & 48.71& 62.39& 42.89\\
Mixtral-8×7b inst & 51.08& 67.38& 49.57\\
GPT-3.5 & \underline{68.32}& \underline{73.13}& \underline{61.85}\\
GPT-4 & \textbf{74.57}& \textbf{91.36}& \textbf{76.29}\\
\midrule
Average score & 37.95& 49.24& 39.48\\
\bottomrule
\end{tabular}
\caption{Code Generation Task Performance}
\label{tab:code_generation}
\end{table}
\begin{table}[htbp]
\centering
\begin{tabular}{lccc}
\toprule
            \textbf{Model} &  \textbf{Generation} &  \textbf{Critiquing} &  \textbf{Correction} \\
\midrule
    Baseline &       - &       40.51 &       65.96 \\
\midrule
Phi-2 & 67.02& 5.94& 66.31\\
LLaMa-2-7b & 47.16& 30.97& 56.38\\
LLaMa-2-70b chat & 47.52& 42.32& 29.08\\
Vicuna-7b & 41.84& 2.04& 64.89\\
Mistral-7b & 62.77& 16.81& 64.89\\
LLaMa-2-13b & 48.58& 5.83& 66.31\\
LLaMa-2-70b chat & 53.90& 44.31& 56.38\\
Vicuna-13b & 61.70& 7.55& 64.54\\
Vicuna-33b & 59.22& 7.77& 66.67\\
LLaMa-2-70b & 77.30& 16.98& 67.02\\
LLaMa-2-70b chat & 62.06& \underline{50.29}& 58.87\\
Mixtral-8×7b & 81.21& 21.43& \underline{68.09}\\
Mixtral-8×7b inst & 84.04& 40.30& 65.96\\
GPT-3.5 & \underline{90.43}& 46.15& 58.16\\
GPT-4 & \textbf{94.68}& \textbf{63.51}& \textbf{77.66}\\
\midrule
Average score & 65.30& 26.81& 62.08\\
\bottomrule
\end{tabular}
\caption{Algorithmic Task Performance}
\label{tab:algorithmic_task}
\end{table}

\newpage
\begin{figure*}
    \centering
    \includegraphics[width=0.9\linewidth]{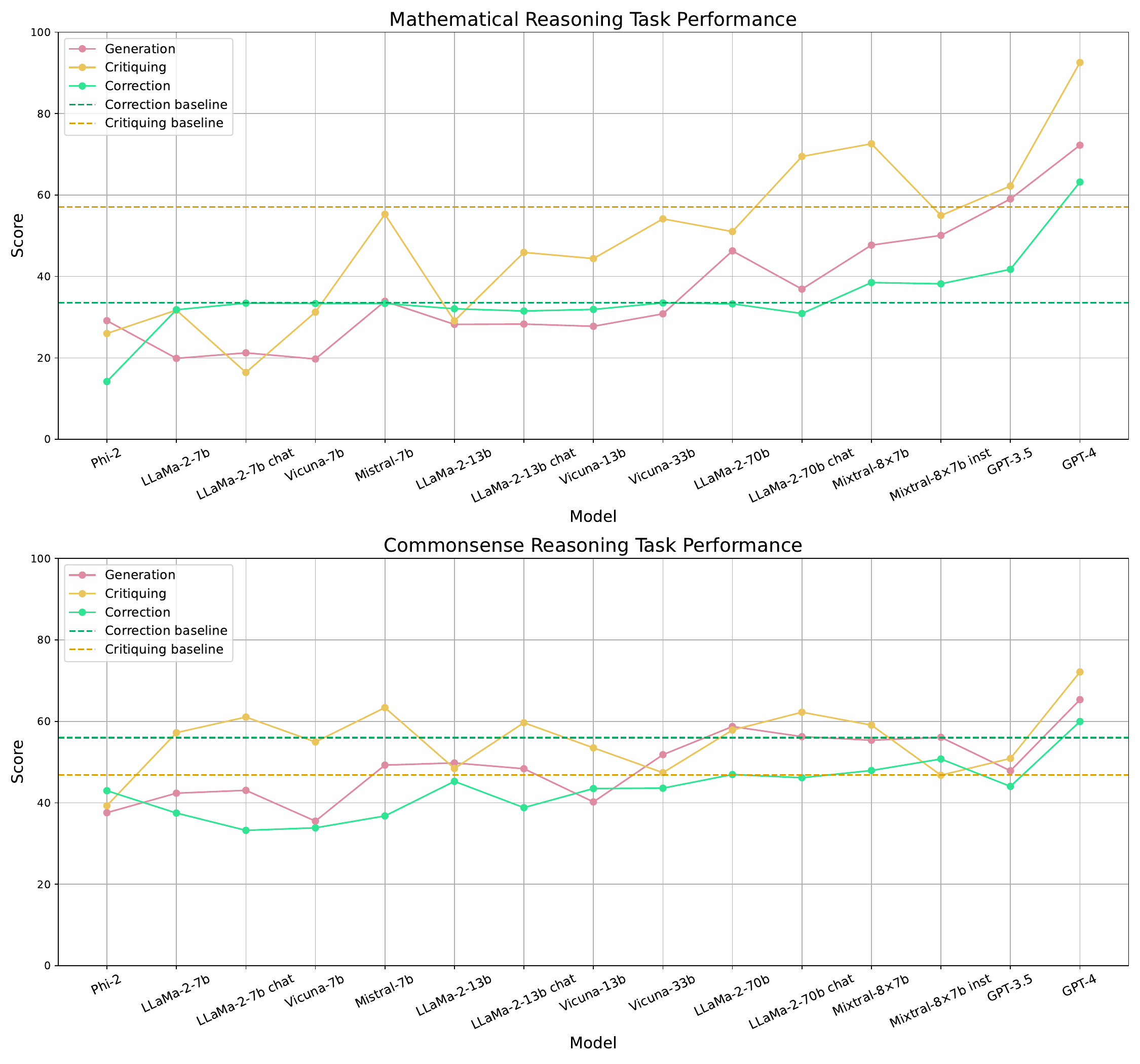}
    \caption{Models' performance on different types of tasks (Part 1).}
    \label{fig:plot_on_diff_type_0}
\end{figure*}
\begin{figure*}
    \centering
    \includegraphics[width=0.9\linewidth]{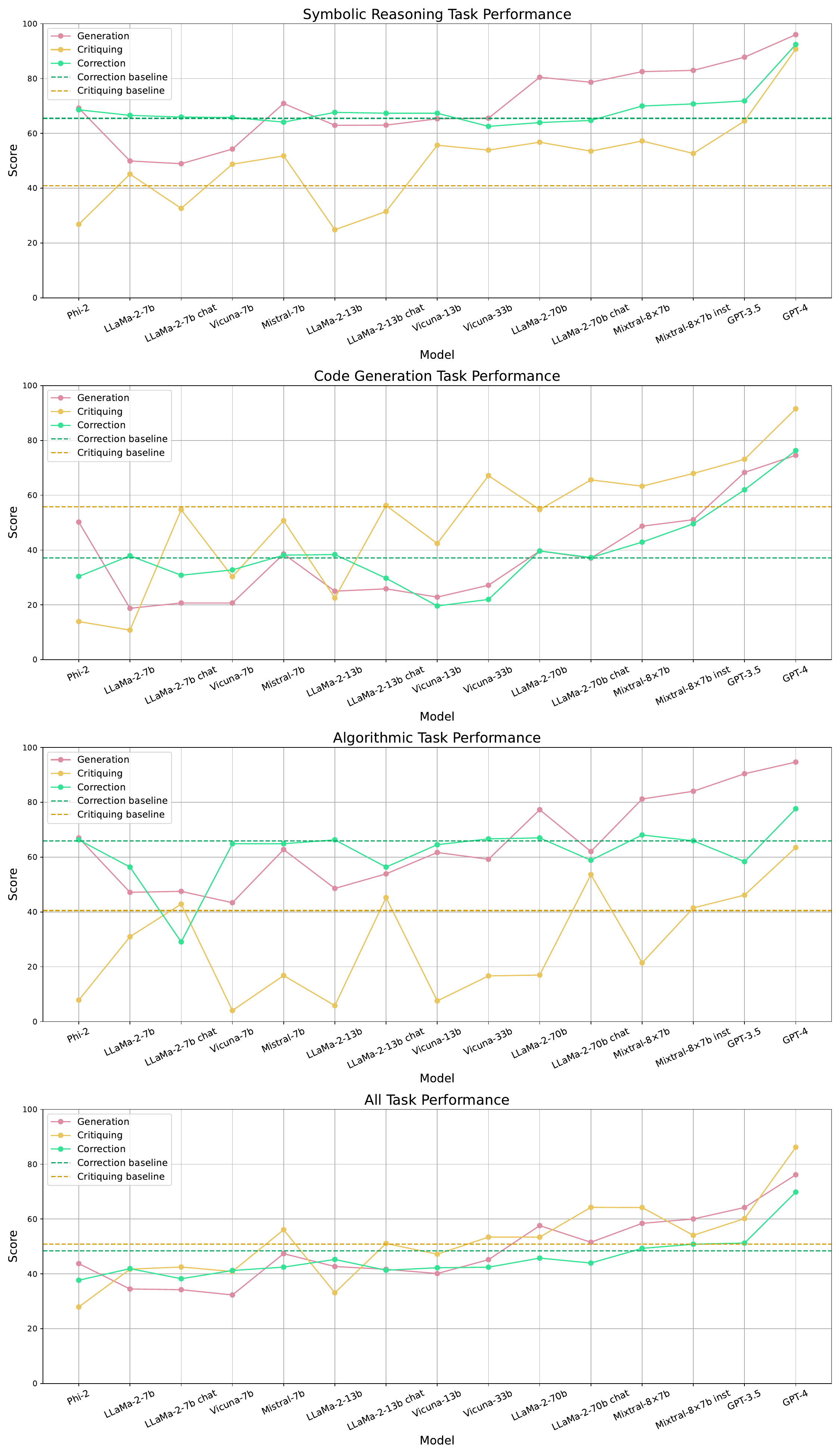}
    \caption{Models' performance on different types of tasks (Part 2).}
    \label{fig:plot_on_diff_type_1}
\end{figure*}
\clearpage

\newpage
\section{Prompt}
\label{appx:prompt}
\subsection{Generation prompt}
In Table \ref{tab:dataset-prompts}, we detailed all the prompts for generation.
\begin{table*}[htbp]
\centering
\begin{tabular}{@{}ll@{}}
\toprule
Dataset         & Prompt Source                 \\ 
\midrule
GSM8K           & CoT \cite{wei2022chain}       \\
MATH            & CoT Hub \cite{fu2023chain}    \\
AQuA            & Self-Consistency \cite{wang2022self}\\
CSQA   & CoT \cite{wei2022chain}        \\
AmbigNQ         & CRITIC \cite{gou2024critic}    \\
StrategyQA      & CoT \cite{wei2022chain}        \\
HotpotQA        & Self-Consistency \cite{wang2022self}  \\
Penguins        & BIG-Bench Hard \cite{suzgun2022challenging}  \\
Colored Object  & BIG-Bench Hard \cite{suzgun2022challenging}  \\
Date            & BIG-Bench Hard \cite{suzgun2022challenging}  \\
MBPP            & DeepSeek-Coder \cite{guo2024deepseek}  \\
Object Counting & BIG-Bench Hard \cite{suzgun2022challenging}  \\
Repeat Copy     & ART \cite{paranjape2023art}  \\
TabMWP          & Written by Authors \\
HumanEval       & Written by Authors  \\
\bottomrule
\end{tabular}
\caption{Generation Prompts and Sources}
\label{tab:dataset-prompts}
\end{table*}

\subsection{Critique prompt}
\begin{table*}[htbp]
\centering
\begin{tabular}{p{\textwidth}}
\toprule
Question: \{question\}\\
Answer: \{response\}\\
Review your previous answer and find problems with your answer.\\
Finally, conclude with either '[[correct]]' if the above answer is correct or '[[wrong]]' if it is incorrect.\\
Verdict: \\
\bottomrule
\end{tabular}
\caption{Zero-shot Answer Only Prompt for Critique}
\end{table*}
\begin{table*}[htbp]
\centering
\resizebox{\linewidth}{!}{
\begin{tabular}{p{\textwidth}}
\toprule
Question: \{question\} \\
Answer: \{response\} \\
Review your previous answer and find problems with your answer. \\
Provide your analysis and verdict in JSON format as follows: \{"analysis": "your analysis", "verdict": "correct or wrong"\}. \\
Analysis and verdict:  \\
\bottomrule
\end{tabular}
}
\caption{Zero-shot Chain of Thought Prompt for Critique}
\end{table*}

\begin{table*}[htbp]
\centering
\resizebox{\linewidth}{!}{%
\begin{tabular}{p{\textwidth}}
\toprule
Review the above answer and determine whether it’s correct. If wrong, find the problems with it.\\
Finally, conclude with either '[[correct]]' if the above answer is correct or '[[wrong]]' if it is incorrect.\\
Here are some examples:\\
---\\
\\
Question: To locate a choker not located in a jewelry box or boutique where would you go? Answer Choices: (a) jewelry store (b) neck (c) jewlery box (d) jewelry box (e) boutique\\
Answer: If you want to locate a choker that is not in a jewelry box or boutique, you would most likely go to your neck (b) since a choker is a type of necklace that is worn around the neck. So the answer is (b).\\
Verdict: The question asks where you would find a choker if it's not located in a jewelry box or boutique. The given answer suggests that the choker would be found on your neck, which is a logical place for a choker when it is being worn.
However, the question seems to be seeking a location where one might find a choker for purchase or retrieval other than a jewelry box or boutique. The options provided are (a) jewelry store, (b) neck, (c) jewelry box, (d) jewelry box, and (e) boutique.\\
Considering the context, option (a) jewelry store is the most appropriate answer. A jewelry store is a common place to find chokers for purchase or display, apart from a jewelry box or boutique. The given answer assumes the choker is being worn, which is not necessarily implied by the question.
So the verdict is: [[wrong]]\\
\\
---\\
......\\
---\\
\\
Question: Stanley had a dream that was very vivid and scary. He had trouble telling it from what? Answer Choices: (a) reality (b) imagination (c) dreamworker (d) nightmare (e) awake\\
Answer: The answer should be something that Stanley had trouble distinguishing his dream from. In this case, the answer is (a) reality. Stanley had trouble telling his vivid and scary dream from reality. So the answer is (a).\\
Verdict: The question asks what Stanley had trouble distinguishing his vivid and scary dream from. The given answer states that the correct choice is (a) reality, indicating that Stanley had difficulty telling his dream apart from real life. This interpretation is consistent with the typical use of the phrase "had trouble telling it from" in the context of dreams. It suggests that the dream felt so real to Stanley that he struggled to differentiate it from actual events or experiences.
So the verdict is: [[correct]]\\
\\
---\\
\\
Please consider the example provided and think it step by step.\\
Question: \{question\}\\
Answer: \{response\}\\
Verdict:\\
\bottomrule
\end{tabular}
}
\caption{Few-shot Prompt (4-shot) of CSQA for Critique}
\end{table*}
\begin{table*}[htbp]
\centering
\resizebox{\linewidth}{!}{%
\begin{tabular}{p{\textwidth}}
\toprule
Review the above answer and determine whether it’s correct. If wrong, find the problems with it.\\
Finally, conclude with either '[[correct]]' if the above answer is correct or '[[wrong]]' if it is incorrect.\\
Here are some examples:\\
---\\
\\
Question: Based on when the project was being worked on, when did the Manhattan project begin and end?\\
Answer: The Manhattan Project began in 1939 and ended in 1946. It was a research and development project during World War II that produced the first nuclear weapons. So the answer is: Began 1939, end 1946\\
Verdict: The Manhattan Project actually began in 1942, not 1939, and was completed in 1946. It was indeed a research and development project during World War II that led to the creation of the first nuclear weapons. The incorrect start date in the answer makes it inaccurate. So the verdict is: [[wrong]]\\
\\
---\\
\\
Question: Who helped lewis and clark on their journey?\\
Answer: Lewis and Clark were helped on their journey by the Corps of Discovery, which included members like Sacagawea, who served as an interpreter and guide. So the answer is: Sacagawea\\
Verdict: The answer to the question ""Who helped Lewis and Clark on their journey?"" is correct. Lewis and Clark were assisted by the Corps of Discovery, which included Sacagawea, an important member who served as an interpreter and guide. Therefore, the answer identifying Sacagawea as a key helper is accurate. So the verdict is: [[correct]]\\
\\
---\\
\\
Question: Who does the voice of cortana in halo video games?\\
Answer: The voice of Cortana in the Halo video games is provided by actress Jen Taylor. So the answer is: Jen Taylor\\
Verdict: The answer provided for the voice of Cortana in the Halo video games is correct. Jen Taylor is indeed the actress who voices Cortana in the series. So the verdict is: [[correct]]\\
\\
---\\
\\
Question: Who is lin's father in legend of korra?\\
Answer: In the animated series ""The Legend of Korra,"" Lin Beifong's father is Toph Beifong, who is a prominent character in the Avatar universe and one of the original members of Team Avatar from the preceding series, ""Avatar: The Last Airbender."" So the answer is: Toph Beifong\\
Verdict: In ""The Legend of Korra,"" Lin Beifong's father is not Toph Beifong. Toph Beifong is Lin's mother. Lin's father is revealed to be a man named Kanto. This information is disclosed in season 4 of the series, making the original answer incorrect. So the verdict is: [[wrong]]\\
\\
---\\
\\
Please consider the example provided and think it step by step.\\
Question: \{question\}\\
Answer: \{response\}\\
Verdict:\\
\bottomrule
\end{tabular}
}
\caption{Few-shot Prompt (4-shot) of AmbigNQ for Critique}
\end{table*}

\begin{table*}[htbp]
\centering
\resizebox{\linewidth}{!}{%
\begin{tabular}{p{\textwidth}}
\toprule
Review the above answer and determine whether it’s correct. If wrong, find the problems with it.\\
Finally, conclude with either '[[correct]]' if the above answer is correct or '[[wrong]]' if it is incorrect.\\
Here are some examples:\\
---\\
\\
Question: Herman likes to feed the birds in December, January and February.  He feeds them 1/2 cup in the morning and 1/2 cup in the afternoon.  How many cups of food will he need for all three months?\\
Answer: Herman feeds the birds 1/2 cup in the morning and 1/2 cup in the afternoon, so each day he uses 1/2 + 1/2 = 1 cup of food. There are 3 months in total, so he will need 3 * 1 = 3 cups of food for all three months. The answer is 3.\\
Verdict: The answer calculates Herman feeding the birds 1 cup of food per day (1/2 cup in the morning and 1/2 cup in the afternoon). However, it incorrectly multiplies this daily amount by 3, assuming there are only 3 days in three months, resulting in 3 cups of food for all three months. This is a significant error in understanding the duration of three months.\\
So the verdict is: [[wrong]]\\
\\
---\\
......\\
---\\
\\
Question: Weng earns \$12 an hour for babysitting. Yesterday, she just did 50 minutes of babysitting. How much did she earn?\\
Answer: Weng earns \$12 per hour for babysitting. Yesterday, she babysat for 50 minutes, which is equivalent to 50/60 = 5/6 of an hour. To calculate her earnings, multiply her hourly rate by the fraction of an hour she worked: $12 * 5/6 = $10. Weng earned \$10 for her 50 minutes of babysitting. The answer is 10.\\
Verdict: The given answer accurately calculates Weng's earnings for babysitting for 50 minutes. It correctly identifies that 50 minutes is 5/6 of an hour and then multiplies this fraction by her hourly rate of \$12. The calculation of \$12 * 5/6 equals \$10, which is the correct amount Weng earned for her 50 minutes of babysitting.\\
So the verdict is: [[correct]]
\\
---\\
\\
Please consider the example provided and think it step by step.\\
Question: \{question\}\\
Answer: \{response\}\\
Verdict:\\
\bottomrule
\end{tabular}
}
\caption{Few-shot Prompt (4-shot) of GSM8K for Critique}
\end{table*}
\begin{table*}[htbp]
\centering
\resizebox{\linewidth}{!}{%
\begin{tabular}{p{\textwidth}}
\toprule
Review the above answer and determine whether it’s correct. If wrong, find the problems with it.\\
Finally, conclude with either '[[correct]]' if the above answer is correct or '[[wrong]]' if it is incorrect.\\
Here are some examples:\\
---\\
......\\
---\\
\\
Table:\\
$[$TITLE$]$: Ferry fares\\
Ferry | Car | Bicycle\\
Seattle-Bremerton | \$12 | \$8\\
Mukilteu-Clinton | \$7 | \$5\\
Southport-Fort Fisher | \$5 | \$2\\
Fauntleroy-Vashon | \$15 | \$5\\
Ocracoke | \$15 | \$3\\
Question: For an economics project, Layla determined the cost of ferry rides for bicycles and cars. How much does it cost to take a bicycle on the Southport-Fort Fisher ferry?
Answer: To find the cost of taking a bicycle on the Southport-Fort Fisher ferry, you can refer to the table:\\
Southport-Fort Fisher | \$5 | \$2\\
The cost for a bicycle on the Southport-Fort Fisher ferry is \$2.\\
So, it costs \$2 to take a bicycle on this ferry.
The answer is 2.\\
Verdict:\\
Upon reviewing the provided answer:\\
The question asks for the cost of taking a bicycle on the Southport-Fort Fisher ferry.
The table provides the cost of ferry rides for both cars and bicycles.\\
The entry for the Southport-Fort Fisher ferry shows the cost for a bicycle as \$2.\\
The answer correctly identifies this value from the table and states that it costs \$2 to take a bicycle on the Southport-Fort Fisher ferry.
No calculations are needed as the information is directly available in the table.\\
The answer accurately extracts the relevant data from the table and correctly identifies the cost of taking a bicycle on the Southport-Fort Fisher ferry as \$2.\\
So the verdict is: [[correct]]\\
\\
---\\
\\
Please consider the example provided and think it step by step.\\
Question: \{question\}\\
Answer: \{response\}\\
Verdict:\\
\bottomrule
\end{tabular}
}
\caption{Few-shot Prompt (4-shot) of TabMWP for Critique}
\end{table*}
\begin{table*}[htbp]
\centering
\begin{adjustbox}{width=\textwidth}
\begin{tabular}{p{\textwidth}}
\toprule
Review the provided code and determine whether it’s correct. If wrong, find the problems with it.\\
Finally, conclude with either '[[correct]]' if the provided answer is correct or '[[wrong]]' if it is incorrect.\\
Here are some examples:\\
---\\
\\
Code:
\begin{lstlisting}[style=Python]
def last_digit_sum(a: int, b: int) -> int:
    """
    Complete the function that takes two integers and returns the sum of their last digits.
    Assume the input is always valid.

    Examples:
    last_digit_sum(123, 453) should return 6 (3 + 3).
    last_digit_sum(789, 10) should return 9 (9 + 0).
    last_digit_sum(100, 27) should return 7 (0 + 7).
    last_digit_sum(25, -31) should return 6 (5 + 1).
    """
    return abs(a \% 100) + abs(b \% 100)
\end{lstlisting}
Verdict: The provided function last\_digit\_sum is supposed to return the sum of the last digits of two integers, but the current implementation returns the sum of the absolute values of the last two digits of each integer. The examples in the problem statement also indicate that the function should return the actual sum of the last digits, not their absolute values.\\
So the verdict is: [[wrong]]\\
\\
---\\
...\\
---\\
\\
Code:
\begin{lstlisting}[style=Python]
from typing import List

def has_similar_word_lengths(words: List[str], threshold: int) -> bool:
    """
    Check if in the given list of words, are any two words whose length difference is less than or equal to the given threshold.
    >>> has_similar_word_lengths(["hello", "world", "hi", "python"], 2)
    True
    >>> has_similar_word_lengths(["chat", "learning", "ai"], 1)
    False
    """
    for i in range(len(words)):
        for j in range(i + 1, len(words)):
            if abs(len(words[i]) - len(words[j])) <= threshold:
                return True
    return False
\end{lstlisting}
Verdict: The provided code iterates over all unique pairs of words by using two nested loops, compares the lengths of the words, and returns True if it finds a pair that satisfies the condition (length difference is less than or equal to the threshold). If no such pair is found, it returns False.\\
So the verdict is: [[correct]]\\
\\
---\\
\\
Please consider the example provided and think it step by step.\\
Code:\\
\{code\}\\
Verdict:\\
\bottomrule
\end{tabular}
\end{adjustbox}
\caption{Few-shot Prompt (4-shot) of HumanEval for Critique}
\end{table*}
\begin{table*}
\centering
\resizebox{\linewidth}{!}{%
\begin{tabular}{p{\textwidth}}
\toprule
Review the provided answer and determine whether it’s correct. If wrong, find the problems with it.\\
Finally, conclude with either '[[correct]]' if the provided answer is correct or '[[wrong]]' if it is incorrect.\\
Here are some examples:\\
---\\
\\
Question: Say pickup a pound of green beans twice, replacing a pound with a bunch for even times and a handful for odd\\
Answer: pickup a handful of green beans pickup a bunch of green beans\\
Verdict: Let's break down the instruction:\\
The instruction is to "pick up a pound of green beans twice".\\
For even times, replace "a pound" with "a bunch".\\
For odd times, replace "a pound" with "a handful".\\
Following these instructions:\\
The first time (odd), "pick up a pound of green beans" becomes "pick up a handful of green beans".
The second time (even), "pick up a pound of green beans" becomes "pick up a bunch of green beans".
Therefore, the provided answer of "pick up a handful of green beans pick up a bunch of green beans" matches the instructions given.\\
So the verdict is: [[correct]]\\
\\
---\\
...\\
---\\
\\
Question: Say stack ten bricks three times, replacing ten with twenty for even times and five for odd times.\\
Answer: stack five bricks stack twenty bricks stack ten bricks.\\
Verdict: According to the instruction:\\
Say "stack ten bricks" three times.\\
Replace "ten" with "twenty" for even times.\\
Replace "ten" with "five" for odd times.\\
Applying these rules:\\
First time (odd): Replace "ten" with "five", resulting in "stack five bricks".\\
Second time (even): Replace "ten" with "twenty", resulting in "stack twenty bricks".\\
Third time (odd again): Replace "ten" with "five", resulting in "stack five bricks".\\
The provided answer incorrectly says "stack ten bricks" for the third time instead of replacing "ten" with "five".
So the verdict is: [[wrong]]\\
\\
---\\
\\
Please consider the example provided and think it step by step.\\
Question: \{question\}\\
Answer: \{solution\}\\
Verdict:\\
\bottomrule
\end{tabular}
}
\caption{Few-shot Prompt (4-shot) of Repeat Copy for Critique}
\end{table*}

\newpage
\subsection{Correction prompt}
\begin{table*}[htbp]
\centering
\resizebox{\linewidth}{!}{
\begin{tabular}{p{\textwidth}}
\toprule
Question: \{question\}\\
Answer: \{answer\}\\
Analysis and verdict: \{critique\}\\
Based on the problems you found, improve your answer. Please reiterate your solution, with your final answer as either a single numerical number, a single word, or a short phrase, in the JSON format as follows: \{"solution":"improved solution", "answer":"your answer"\}.\\
Improved solution and answer:\\
\bottomrule
\end{tabular}
}
\caption{Zero-shot Chain of Thought Prompt for Correction}
\end{table*}
\begin{table*}[htbp]
\centering
\resizebox{\linewidth}{!}{%
\begin{tabular}{p{\textwidth}}
\toprule
Based on the problems you found, improve your answer. You must choose only one option from A to E. Please reiterate your answer, with your final answer a single letter from A to E, in the form [[answer]].\\
Here are some examples:\\
---\\
\\
Question: To locate a choker not located in a jewelry box or boutique where would you go? Answer Choices: (a) jewelry store (b) neck (c) jewlery box (d) jewelry box (e) boutique\\
Answer: If you want to locate a choker that is not in a jewelry box or boutique, you would most likely go to your neck (b) since a choker is a type of necklace that is worn around the neck. So the answer is (b).
Verdict: The question asks where you would find a choker if it's not located in a jewelry box or boutique. The given answer suggests that the choker would be found on your neck, which is a logical place for a choker when it is being worn.\\
However, the question seems to be seeking a location where one might find a choker for purchase or retrieval other than a jewelry box or boutique. The options provided are (a) jewelry store, (b) neck, (c) jewelry box, (d) jewelry box, and (e) boutique.\\
Considering the context, option (a) jewelry store is the most appropriate answer. A jewelry store is a common place to find chokers for purchase or display, apart from a jewelry box or boutique. The given answer assumes the choker is being worn, which is not necessarily implied by the question.\\
So the verdict is: [[wrong]]\\
Improved Answer: The improved answer is: [[a]] jewelry store.\\
\\
---\\
......\\
---\\
\\
Question: Stanley had a dream that was very vivid and scary. He had trouble telling it from what? Answer Choices: (a) reality (b) imagination (c) dreamworker (d) nightmare (e) awake\\
Answer: The answer should be something that Stanley had trouble distinguishing his dream from. In this case, the answer is (a) reality. Stanley had trouble telling his vivid and scary dream from reality. So the answer is (a).\\
Verdict: The question asks what Stanley had trouble distinguishing his vivid and scary dream from. The given answer states that the correct choice is (a) reality, indicating that Stanley had difficulty telling his dream apart from real life. This interpretation is consistent with the typical use of the phrase "had trouble telling it from" in the context of dreams. It suggests that the dream felt so real to Stanley that he struggled to differentiate it from actual events or experiences.
So the verdict is: [[correct]]\\
Improved Answer: The answer is [[a]]\\
\\
---\\
\\
Question: \{question\}\\
Answer: \{answer\}\\
Verdict: \{critique\}\\
Improved Answer:\\
\bottomrule
\end{tabular}
}
\caption{Few-shot Prompt (4-shot) of CSQA for Correction}
\end{table*}
\begin{table*}[htbp]
\centering
\resizebox{\linewidth}{!}{%
\begin{tabular}{p{\textwidth}}
\toprule
Based on the problems you found, improve your answer. Please reiterate your answer, with your final answer as either a single numerical number, a single word, or a short phrase, in the form [[answer]].\\
Here are some examples:\\
---\\
\\
Question: Based on when the project was being worked on, when did the Manhattan project begin and end?\\
Answer: The Manhattan Project began in 1939 and ended in 1946. It was a research and development project during World War II that produced the first nuclear weapons. So the answer is: Began 1939, end 1946\\
Verdict: The Manhattan Project actually began in 1942, not 1939, and was completed in 1946. It was indeed a research and development project during World War II that led to the creation of the first nuclear weapons. The incorrect start date in the answer makes it inaccurate. So the verdict is: [[wrong]]\\
Improved Answer: The Manhattan Project [[began in 1942 and ended in 1946]].\\
\\
---\\
\\
Question: Who helped lewis and clark on their journey?\\
Answer: Lewis and Clark were helped on their journey by the Corps of Discovery, which included members like Sacagawea, who served as an interpreter and guide. So the answer is: Sacagawea\\
Verdict: The answer to the question "Who helped Lewis and Clark on their journey?" is correct. Lewis and Clark were assisted by the Corps of Discovery, which included Sacagawea, an important member who served as an interpreter and guide. Therefore, the answer identifying Sacagawea as a key helper is accurate. So the verdict is: [[correct]]\\
Improved Answer: [[Sacagawea]] helped Lewis and Clark on their journey.\\
\\
---\\
\\
Question: Who does the voice of cortana in halo video games?\\
Answer: The voice of Cortana in the Halo video games is provided by actress Jen Taylor. So the answer is: Jen Taylor\\
Verdict: The answer provided for the voice of Cortana in the Halo video games is correct. Jen Taylor is indeed the actress who voices Cortana in the series. So the verdict is: [[correct]]\\
Improved Answer: [[Jen Taylor]]\\
\\
---\\
\\
Question: Who is lin's father in legend of korra?\\
Answer: In the animated series "The Legend of Korra," Lin Beifong's father is Toph Beifong, who is a prominent character in the Avatar universe and one of the original members of Team Avatar from the preceding series, "Avatar: The Last Airbender." So the answer is: Toph Beifong\\
Verdict: In "The Legend of Korra," Lin Beifong's father is not Toph Beifong. Toph Beifong is Lin's mother. Lin's father is revealed to be a man named Kanto. This information is disclosed in season 4 of the series, making the original answer incorrect. So the verdict is: [[wrong]]
Improved Answer: Lin Beifong's father in "The Legend of Korra" is [[Kanto]].\\
\\
---\\
\\
Question: \{question\}\\
Answer: \{answer\}\\
Verdict: \{critique\}\\
Improved Answer:\\
\bottomrule
\end{tabular}
}
\caption{Few-shot Prompt (4-shot) of AmbigNQ for Correction}
\end{table*}
\begin{table*}[htbp]
\centering
\resizebox{\linewidth}{!}{%
\begin{tabular}{p{\textwidth}}
\toprule
Based on the problems you found, improve your answer. Please reiterate your answer, with your final answer a single numerical number, in the form [[answer]].\\
Here are some examples:\\
---\\
\\
Question: Herman likes to feed the birds in December, January and February.  He feeds them 1/2 cup in the morning and 1/2 cup in the afternoon.  How many cups of food will he need for all three months?
Answer: Herman feeds the birds 1/2 cup in the morning and 1/2 cup in the afternoon, so each day he uses 1/2 + 1/2 = 1 cup of food. There are 3 months in total, so he will need 3 * 1 = 3 cups of food for all three months. The answer is 3.\\
Verdict: The answer calculates Herman feeding the birds 1 cup of food per day (1/2 cup in the morning and 1/2 cup in the afternoon). However, it incorrectly multiplies this daily amount by 3, assuming there are only 3 days in three months, resulting in 3 cups of food for all three months. This is a significant error in understanding the duration of three months.
So the verdict is: [[wrong]]\\
Improved Answer: To calculate the amount of food Herman will need for all three months, we need to consider the number of days in December, January, and February. December and January typically have 31 days each, while February can have 28 or 29 days, depending on whether it's a leap year. Assuming a non-leap year, February has 28 days. Therefore, the total number of days is 31 (December) + 31 (January) + 28 (February) = 90 days. Since Herman feeds the birds 1 cup of food each day (1/2 cup in the morning and 1/2 cup in the afternoon), the total amount of food he will need for all three months is 90 cups. The answer is [[90]].\\
\\
---\\
......\\
---\\
\\
Question: Weng earns \$12 an hour for babysitting. Yesterday, she just did 50 minutes of babysitting. How much did she earn?\\
Answer: Weng earns \$12 per hour for babysitting. Yesterday, she babysat for 50 minutes, which is equivalent to 50/60 = 5/6 of an hour. To calculate her earnings, multiply her hourly rate by the fraction of an hour she worked: \$12 * 5/6 = \$10. Weng earned \$10 for her 50 minutes of babysitting. The answer is 10.\\
Verdict: The given answer accurately calculates Weng's earnings for babysitting for 50 minutes. It correctly identifies that 50 minutes is 5/6 of an hour and then multiplies this fraction by her hourly rate of \$12. The calculation of \$12 * 5/6 equals \$10, which is the correct amount Weng earned for her 50 minutes of babysitting.\\
So the verdict is: [[correct]]\\
Improved Answer: The answer is [[10]].\\
\\
---\\
\\
Question: \{question\}\\
Answer: \{answer\}\\
Verdict: \{critique\}\\
Improved Answer:\\
\bottomrule
\end{tabular}
}
\caption{Few-shot Prompt (4-shot) of GSM8K for Correction}
\end{table*}
\begin{table*}
\centering
\begin{adjustbox}{width=\textwidth}
\begin{tabular}{p{\textwidth}}
\toprule
Based on the problems you found, improve your answer. Please reiterate your answer, with your final answer as either a single numerical number or one option from A to D, in the form [[answer]].\\
Here are some examples:\\
---\\
......\\
---\\
\\
Table:\\
$[$TITLE$]$: Ferry fares\\
Ferry | Car | Bicycle\\
Seattle-Bremerton | \$12 | \$8\\
Mukilteu-Clinton | \$7 | \$5\\
Southport-Fort Fisher | \$5 | \$2\\
Fauntleroy-Vashon | \$15 | \$5\\
Ocracoke | \$15 | \$3\\
Question: For an economics project, Layla determined the cost of ferry rides for bicycles and cars. How much does it cost to take a bicycle on the Southport-Fort Fisher ferry?
Answer: To find the cost of taking a bicycle on the Southport-Fort Fisher ferry, you can refer to the table:\\
Southport-Fort Fisher | \$5 | \$2\\
The cost for a bicycle on the Southport-Fort Fisher ferry is \$2.\\
So, it costs \$2 to take a bicycle on this ferry.
The answer is 2.\\
Verdict:\\
Upon reviewing the provided answer:\\
The question asks for the cost of taking a bicycle on the Southport-Fort Fisher ferry.
The table provides the cost of ferry rides for both cars and bicycles.\\
The entry for the Southport-Fort Fisher ferry shows the cost for a bicycle as \$2.\\
The answer correctly identifies this value from the table and states that it costs \$2 to take a bicycle on the Southport-Fort Fisher ferry.
No calculations are needed as the information is directly available in the table.\\
The answer accurately extracts the relevant data from the table and correctly identifies the cost of taking a bicycle on the Southport-Fort Fisher ferry as \$2.\\
So the verdict is: [[correct]]\\
Improved Answer: The answer is [[2]].\\
\\
---\\
\\
Question: \{question\}\\
Answer: \{answer\}\\
Verdict: \{critique\}\\
Improved Answer:\\
\bottomrule
\end{tabular}
\end{adjustbox}
\caption{Few-shot Prompt (4-shot) of TabMWP for Correction}
\end{table*}
\begin{table*}
\centering
\begin{adjustbox}{width=\textwidth}
\begin{tabular}{p{\textwidth}}
\toprule
Based on the problems you found, improve your code in the form [BEGIN]\textbackslash nimproved code\textbackslash n[DONE]
Here are some examples:\\
---\\
......\\
---\\
Code:
\begin{lstlisting}[style=Python]
from typing import List

def has_similar_word_lengths(words: List[str], threshold: int) -> bool:
    """
    Check if in the given list of words, are any two words whose length difference is less than or equal to the given threshold.
    >>> has_similar_word_lengths(["hello", "world", "hi", "python"], 2)
    True
    >>> has_similar_word_lengths(["chat", "learning", "ai"], 1)
    False
    """
    for i in range(len(words)):
        for j in range(i + 1, len(words)):
            if abs(len(words[i]) - len(words[j])) <= threshold:
                return True
    return False
\end{lstlisting}
Verdict: The provided code iterates over all unique pairs of words by using two nested loops, compares the lengths of the words, and returns True if it finds a pair that satisfies the condition (length difference is less than or equal to the threshold). If no such pair is found, it returns False.\\
So the verdict is: [[correct]]\\
Improved Code:\\
$[$BEGIN$]$
\begin{lstlisting}[style=Python]
from typing import List

def has_similar_word_lengths(words: List[str], threshold: int) -> bool:
    """
    Check if in the given list of words, are any two words whose length difference is less than or equal to the given threshold.
    >>> has_similar_word_lengths(["hello", "world", "hi", "python"], 2)
    True
    >>> has_similar_word_lengths(["chat", "learning", "ai"], 1)
    False
    """
    for i in range(len(words)):
        for j in range(i + 1, len(words)):
            if abs(len(words[i]) - len(words[j])) <= threshold:
                return True
    return False
\end{lstlisting}
$[$DONE$]$\\
\\
---\\
\\
Code:\\
\{code\}\\
Verdict: \{critique\}\\
Improved Code:\\
\bottomrule
\end{tabular}
\end{adjustbox}
\caption{Few-shot Prompt (4-shot) of HumanEval for Correction}
\end{table*}
\begin{table*}
\centering
\resizebox{\linewidth}{!}{%
\begin{tabular}{p{\textwidth}}
\toprule
Based on the problems you found, improve your answer. Please reiterate your answer, with your final answer in the form [[answer]].\\
Here are some examples:\\
---\\
\\
Question: Say pickup a pound of green beans twice, replacing a pound with a bunch for even times and a handful for odd\\
Answer: pickup a handful of green beans pickup a bunch of green beans
Verdict: Let's break down the instruction:\\
The instruction is to "pick up a pound of green beans twice".\\
For even times, replace "a pound" with "a bunch".\\
For odd times, replace "a pound" with "a handful".\\
Following these instructions:\\
The first time (odd), "pick up a pound of green beans" becomes "pick up a handful of green beans".\\
The second time (even), "pick up a pound of green beans" becomes "pick up a bunch of green beans".\\
Therefore, the provided answer of "pick up a handful of green beans pick up a bunch of green beans" matches the instructions given.
So the verdict is: [[correct]]\\
Improved Answer: [[pickup a handful of green beans pickup a bunch of green beans]]\\
\\
---\\
......\\
---\\
\\
Question: Say stack ten bricks three times, replacing ten with twenty for even times and five for odd times.\\
Answer: stack five bricks stack twenty bricks stack ten bricks.\\
Verdict: According to the instruction:\\
Say "stack ten bricks" three times.\\
Replace "ten" with "twenty" for even times.\\
Replace "ten" with "five" for odd times.\\
Applying these rules:\\
First time (odd): Replace "ten" with "five", resulting in "stack five bricks".\\
Second time (even): Replace "ten" with "twenty", resulting in "stack twenty bricks".\\
Third time (odd again): Replace "ten" with "five", resulting in "stack five bricks".\\
The provided answer incorrectly says "stack ten bricks" for the third time instead of replacing "ten" with "five".\\
So the verdict is: [[wrong]]\\
Improved Answer: [[stack five bricks stack twenty bricks stack five bricks]]\\
\\
---\\
\\
Question: \{question\}\\
Answer: \{answer\}\\
Verdict: \{critique\}\\
Improved Answer:\\
\bottomrule
\end{tabular}
}
\caption{Few-shot Prompt (4-shot) of Repeat Copy for Correction}
\end{table*}

\end{document}